\documentclass[times,twocolumn,final]{article}

\usepackage{framed,multirow}

\usepackage{amssymb}
\usepackage{latexsym}
\usepackage{natbib}
\usepackage{graphicx}

\usepackage{url}
\usepackage{xcolor}
\usepackage[nolist]{acronym}
\usepackage{hyperref}
\usepackage{pifont}
\usepackage{amssymb}
\usepackage{subfig}
\usepackage{authblk}
\usepackage{latexsym}
\newcommand{\cmark}{\ding{51}}%
\newcommand{\xmark}{\ding{55}}%

\definecolor{newcolor}{rgb}{.8,.349,.1}

\title{Domain generalization across tumor types, laboratories, and species - insights from the 2022 edition of the Mitosis Domain Generalization Challenge}%

\author[1]{Marc Aubreville}
\author[2]{Nikolas Stathonikos}
\author[3]{Taryn A. Donovan}
\author[4]{Robert Klopfleisch}
\author[1]{Jonas Ammeling}
\author[1]{Jonathan Ganz}
\author[5,6]{Frauke Wilm}
\author[7]{Mitko Veta}
\author[10]{Samir Jabari}
\author[8]{Markus Eckstein}
\author[A]{Jonas Annuscheit}
\author[A]{Christian Krumnow}
\author[B]{Engin Bozaba}
\author[B]{Sercan Çayır}
\author[C]{Hongyan Gu}
\author[C]{Xiang 'Anthony' Chen}
\author[D]{Mostafa Jahanifar}
\author[D]{Adam Shephard}
\author[E1]{Satoshi Kondo}
\author[E2]{Satoshi Kasai}
\author[F]{Sujatha Kotte}
\author[F]{VG Saipradeep}
\author[G]{Maxime W. Lafarge}
\author[G]{Viktor H. Koelzer}
\author[H]{Ziyue Wang}
\author[H]{Yongbing Zhang}
\author[I1]{Sen Yang}
\author[I2]{Xiyue Wang}
\author[6]{Katharina Breininger}
\author[9]{Christof A. Bertram}

\affil[1]{Technische Hochschule Ingolstadt, Ingolstadt, Germany}
\affil[2]{Pathology Department, UMC Utrecht, The Netherlands}
\affil[3]{Department of Anatomic Pathology, The Schwarzman Animal Medical Center, New York, USA}
\affil[4]{Institute of Veterinary Pathology, Freie Universit{\"a}t Berlin, Berlin, Germany}
\affil[5]{Pattern Recognition Lab, Friedrich-Alexander-Universität Erlangen-Nürnberg, Erlangen, Germany}
\affil[6]{Department Artificial Intelligence in Biomedical Engineering, Friedrich-Alexander-Universität Erlangen-Nürnberg, Erlangen, Germany}
\affil[7]{Computational Pathology Group, Radboud UMC Nijmegen, The Netherlands}
\affil[10]{Institute of Neuropathology, University Hospital Erlangen, Friedrich-Alexander-Universität Erlangen-Nürnberg, Erlangen, Germany}
\affil[8]{Institute of Pathology, University Hospital Erlangen, Friedrich-Alexander-Universität Erlangen-Nünberg, Erlangen, Germany}
\affil[A]{University of Applied Sciences (HTW) Berlin, Berlin, Germany}
\affil[B]{Artificial Intelligence Research Team, Virasoft Corporation, New York, USA}
\affil[C]{University of California, Los Angeles, USA}
\affil[D]{University of Warwick, United Kingdom}
\affil[E1]{Muroran Institute of Technology, Muroran, Japan}
\affil[E2]{Niigata University of Health and Welfare, Niigata, Japan}
\affil[F]{TCS Research, Tata Consultancy Services Ltd, Hyderabad, India}
\affil[G]{Department of Pathology and Molecular Pathology, University Hospital Zurich, University of Zurich, Zurich, Switzerland}
\affil[H]{Harbin Institute of Technology, Shenzhen, China}
\affil[I1]{College of Biomedical Engineering, Sichuan University, Chengdu, China}
\affil[I2]{Department of Radiation Oncology, Stanford University School of Medicine, Palo Alto, USA}
\affil[9]{Institute of Pathology, University of Veterinary Medicine, Vienna, Austria}

\begin{document}
\maketitle


\begin{abstract}
Recognition of mitotic figures in histologic tumor specimens is highly relevant to patient outcome assessment. This task is challenging for algorithms and human experts alike, with deterioration of algorithmic performance under shifts in image representations. Considerable covariate shifts occur when assessment is performed on different tumor types, images are acquired using different digitization devices, or specimens are produced in different laboratories. This observation motivated the inception of the 2022 challenge on MItosis Domain Generalization (MIDOG 2022). The challenge provided annotated histologic tumor images from six different domains and evaluated the algorithmic approaches for mitotic figure detection provided by nine challenge participants on ten independent domains. Ground truth for mitotic figure detection was established in two ways: a three-expert { majority vote} and an independent, immunohistochemistry-assisted set of labels. This work represents an overview of the challenge tasks, the algorithmic strategies employed by the participants, and potential factors contributing to their success. With an $F_1$ score of 0.764 for the top-performing team, we summarize that domain generalization across various tumor domains is possible with today's deep learning-based recognition pipelines. { However, we also found that domain characteristics not present in the training set (feline as new species, spindle cell shape as new morphology and a new scanner) led to small but significant decreases in performance.} When assessed against the immunohistochemistry-assisted reference standard, all methods resulted in reduced recall scores, with only minor changes in the order of participants in the ranking. 

%
\end{abstract}


\section{Introduction}

Despite advances in molecular characterization of biological tumor behavior, morphological tumor classification using established histopathologic techniques remains an important factor in tumor prognostication \cite{makki2015diversity,soliman2016ki}. One criterion of particular interest within many tumor grading schemes is the density of cells undergoing division, which are visible as \acp{MF} in \ac{HE}-stained histopathological sections \citep{veta2015assessment,veta2019predicting}. The number of \acp{MF} within a specific tumor area is enumerated by experienced pathologists, resulting in the \ac{MC}. Despite the prognostic relevance of the \ac{MC}, low inter-rater consistency on an object level has been reported in many studies \citep{veta2016mitosis,Meyer:2005cl,Meyer:2009eu,malon2012mitotic,Bertram2021VetPathol}. The recommendation for pathologists is to select the region of the suspected highest mitotic activity, which is considered to be the best predictor of tumor behavior \citep{Azzola:2003ey,Meuten:2008bs,veta2015assessment}. Selection of this \ac{ROI} within the tumor has a great impact on the \ac{MC} \citep{bertram2020computerized}, but is difficult for pathologists to reliably accomplish and is poorly reproducible \citep{aubreville2020deep,Bertram2021VetPathol}. While assessment of mitotic activity in the entire tumor section (or in the case of the digital image: the \ac{WSI}) would be preferable in order to identify those mitotic hotspot \ac{ROI}, this is not feasible in current practice. Additionally, low inter-rater consistency on an object level within these selected \ac{ROI} has been reported in many studies with the tendency of pathologists to overlook \acp{MF} \citep{veta2016mitosis,Meyer:2005cl,Meyer:2009eu,malon2012mitotic,Bertram2021VetPathol}. The combination of these circumstances and the recent availability of large-scale digital pathology solutions makes automatic detection of \acp{MF} desirable. 

\begin{figure*}[t]
\includegraphics[width=\textwidth]{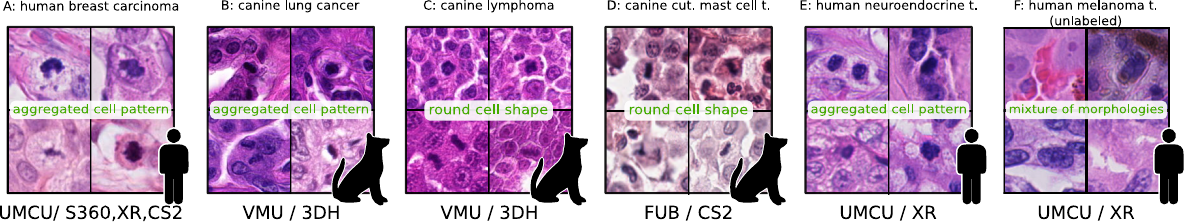}
\caption{ Random selection of crops of size $128\times 128$ px, centered around annotated \acp{MF} from the six domains of the training set. { Caption indicates the originating lab (UMCU = UMC Utrecht, VMU = University of Veterinary Medicine Vienna, FUB = FU of Berlin) and the scanners (S360 = Hamamatsu S360, XR = Hamamatsu XR, CS2 = Aperio ScanScope CS2, 3DH = 3DHIstech Pannoramic Scan II). } Domain F was not labeled, hence the crops were selected at random.}
\label{fig:tumortypes}
\end{figure*}

Unsurprisingly, \ac{MF} detection was one of the earliest identified areas of research interest in computational pathology, with the first approaches in 2008 \citep{malon2008identifying}. The first challenge on \ac{MF} detection in breast cancer (MITOS2012, \citep{LUDOVIC20138}) was held at \ac{ICPR} and resulted in the first publicly available \ac{MF} dataset. While this gave rise to algorithm development in the field, it was also an example of questionable dataset quality, as the training and test sets were selected from the same histology slides \citep{LUDOVIC20138}. More recent challenges (MITOS2014 \citep{roux2014mitos}, AMIDA13 \citep{veta2015assessment}, TUPAC16 \citep{veta2019predicting}) also comprised breast cancer and incorporated a higher number of cases, yet, were still limited by having the same digitization device for the training and test set. 

As shown by prior research \citep{aubreville2021quantifying}, the digitization device has a decisive influence on the detection quality, as it coincides with a shift in the image representation, leading to a domain shift in the latent representation of the detection models \citep{stacke2020measuring,aubreville2023mitosis}. Investigation of these limitations was the main idea behind the \ac{MIDOG} challenge, held as a one-time event at the International Conference on \ac{MICCAI} in 2021. This challenge, which was the first to directly target domain generalization in histopathology, evaluated the detection of \acp{MF} in \acp{ROI} of human breast cancer, digitized using various devices (\ac{WSI} scanners). 

Since \acp{MF} are not only of interest for human breast cancer, the 2022 \ac{MIDOG} challenge extended the task of \ac{MF} domain generalization to include further representation shifts of interest: In addition to the use of different \ac{WSI} scanners, the training dataset was enhanced by including histological specimens from different tumor types as well as different species (human, canine, feline), processed by different laboratories. Each of these contributing factors defined a \textit{tumor domain}. We define a tumor domain as a specific combination of tumor type, species, lab, and \ac{WSI} scanner. We found that the domain gap between tumor types is substantial \citep{aubreville_comprehensive_2023} and seems to be more important than the domain gap between scanners, thus the cases used for the \ac{MIDOG} 2022 challenge were primarily categorized by the tumor { type}.

\subsection*{Challenge format and task}

As in previous challenges on \ac{MF} detection, we provided \acp{ROI}, selected by an experienced pathologist from a tumor region with the presumed highest mitotic activity and appropriate tissue and scan quality. \ac{MF} candidates were identified and assessed by a blinded { majority vote of three experts (with the third expert only asked if the first two disagreed)}. The training set, consisting of 405 tumor cases (corresponding to 405 patients) and featuring 9,501 \ac{MF} annotations was released on April 20, 2022. These cases were split across six tumor domains (see Fig. \ref{fig:tumortypes}), out of which five were provided with labels and one was provided without labels as an additional data source for unsupervised domain adaptation techniques.  An extended version of the training set, including two novel domains, was made available under a Creative Commons CC0 license post-challenge \citep{aubreville_comprehensive_2023}.

The participants were required to package their algorithmic solution in the form of a docker container\footnote{A reference docker container for evaluation was made available to the participants at: \url{https://github.com/DeepMicroscopy/MIDOG_evaluation_docker}}, which was subsequently evaluated on the test data on the grand-challenge.org platform\footnote{\url{https://midog2022.grand-challenge.org}} in a fully automatic manner, i.e., no participant had access to any of the test images during or after the challenge. To perform a technical validation of the docker containers, we provided an independent preliminary test set, consisting of four unseen tumor domains. During a preliminary test phase, which started on August 5, participants were allowed to perform one evaluation of an algorithmic approach per day. We explicitly made the participants aware that the four domains of the preliminary test set were disjointed from the tumor domains of the actual challenge test set, so overfitting to those domains by means of hyperparameter or model selection would not be meaningful. The final challenge submission phase started on August 26 and lasted until August 30. During this phase, participating teams were exclusively authorized to submit a single algorithmic approach.

The challenge provided two tracks: As multiple openly accessible datasets on \ac{MF} detection already exist, we gave participants the choice to either use only data provided by the challenge (track 1) or additionally use publicly available data and labels (track 2). In the second track, participants also had the option to use in-house datasets under the condition that these datasets were made publicly available and announced on the challenge website up to one month prior to the challenge. We opted for this strategy to maximize the reproducibility of the challenge results. However, no participating team chose to use previously non-public datasets.  

The structured challenge design includes details about the policies regarding participation, publication, awards, and results announcement, and was made available publicly \citep{marc_aubreville_2022_6362337}.  The challenge design was proposed and evaluated in a single-blinded peer review for admission to \ac{MICCAI} 2022.  

\subsection*{Main novelties over the predecessor}
While the task (\acp{MF} detection on \acp{ROI} images) was identical to the preceding MIDOG 2021 challenge, we incorporated three major modifications in the 2022 challenge design that set it apart from its predecessor:
\begin{itemize}
    \item We extended the sources of domain shift by not only including the imaging device and the inherent stain differences between cases but also by incorporating different laboratories (and hence tissue processing), different tumor types, and different species, minimizing the gap to real-world data variability.
    \item The evaluation was carried out on ten independent tumor domains, representing a wide variety of conditions and thus allowing for better generalization of the assessment. The ten domains were additionally disjoint from the four independent domains of the preliminary test used for technical validation of the docker pipeline.
    \item We established the ground truth of the test set not only as the { majority vote} of three experts on the \ac{HE}-stained sections (used for challenge evaluation and ranking) but also by additionally using an \ac{IHC} stain for \ac{PHH3} (specific for cells entering the mitotic cycle  \citep{hendzel1997mitosis}), which was superimposed on the \ac{HE} image for assisted labeling aiming to object-level confusion, which is a main source of inter-rater disagreement \citep{veta2016mitosis}.

\end{itemize}

\section{Material and evaluation methods}
For all tumor types included in our datasets, the \ac{MC} has well-established prognostic relevance for discriminating patient outcome, either as a solitary prognostic test or as part of an established grading scheme. We retrieved human tissue samples from the \acp{DA} of the Department of Pathology of the \ac{UMC} Utrecht, The Netherlands, as well as the Institute of Neuropathology and the Institute of Pathology of the University Hospital Erlangen, Germany. All samples were prepared from paraffin-embedded tumor sections stained according to the standard procedures of the respective institutions. We received ethics approval from the UMC Utrecht (TCBio 20-776) and the ethics board of the medical faculty of FAU Erlangen-Nürnberg (AZ 92\_14B, AZ 193\_18B, 22\_342\_B). For samples taken from the \acp{DA} of veterinary
pathology laboratories (\ac{FUB}, Germany and \ac{VMU}, Austria), no ethics approval was required.

\subsection{Challenge cohort and tumor domains}
\begin{figure*}
\includegraphics[width=\textwidth]{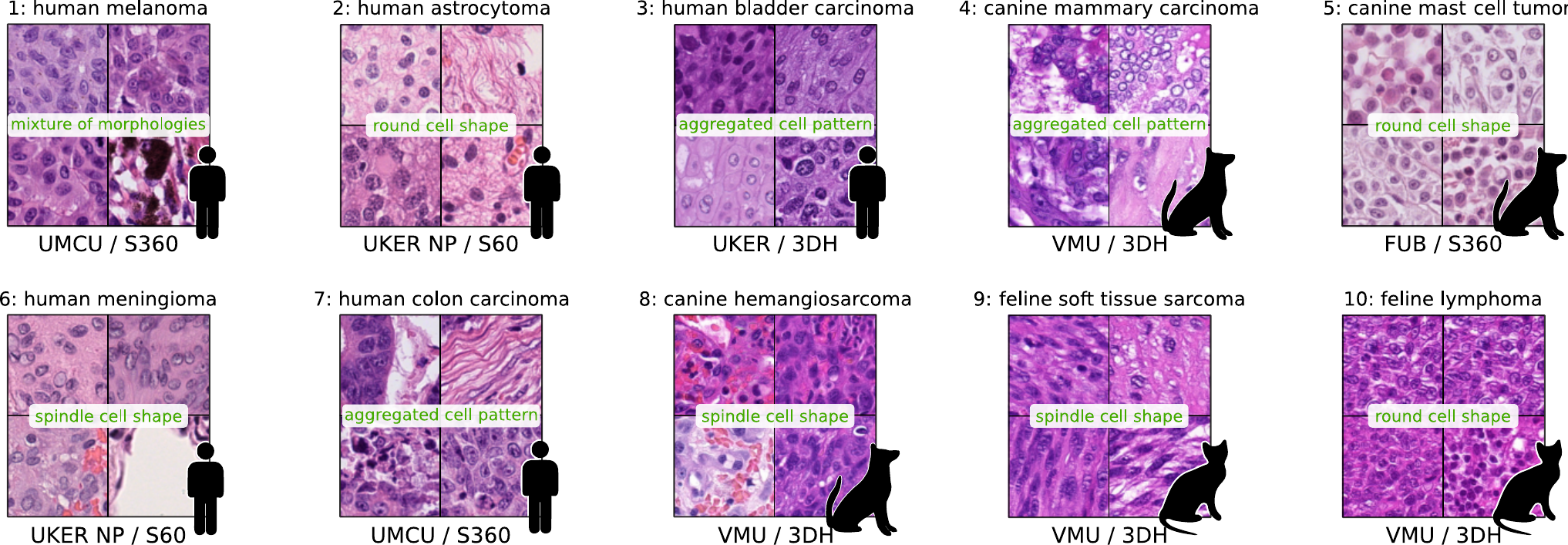}
\caption{  Overview of the domains of the test set. Random cropouts sized $256\times 256$ px from four randomly selected images of each domain are shown. Caption indicates origin of tissue (UMCU = UMC Utrecht, UKER = University Hospital Erlangen, UKER NP = Institute of Neuropathology at University Hospital Erlangen, FUB = FU Berlin, VMU = University of Veterinary Medicine Vienna) and scanner (S360 = Hamamatsu S360, S60 = Hamamatsu S60, 3DH = 3DHistech Pannoramic Scan II). The tumor types are categorized by the tissue morphology into aggregated cell patterns, round cell morphology and spindle cell morphology. } 
\label{fig:overview_domains_testset}
\end{figure*}
{
In our datasets, we included tumors from multiple different morphological categories: aggregated cell pattern, round cell shape, and spindle cell shape. While these categories were used in order to allow comparison of the algorithmic performance depending on the tumor morphology, we acknowledge that some tumor types (see below) are difficult to group into these categories and the best fitting category was chosen.   
In the training dataset, we included 405 cases (see Fig. \ref{fig:tumortypes}), split into the following domains:}

\begin{itemize}
    \item Domain A: Human breast carcinoma{, an epithelial tumor with aggregated cell pattern/morphology}, retrieved from the \ac{DA} of UMC Utrecht. 150 cases split across three scanners (Hamamatsu XR, Hamamatsu S360, Aperio Scanscope CS2, 50 each) at 40$\times$ magnification (0.23 to 0.25 $\mu m$/px), previously released as training set of the 2021 \ac{MIDOG} challenge \citep{aubreville2023mitosis}. The \ac{MC} is part of the College of American Pathologists guidelines for breast cancer \citep{fitzgibbons2023protocol}. 
    \item Domain B: Canine lung carcinoma { an epithelial tumor with aggregated cell pattern/morphology}, retrieved from the \ac{DA} of \ac{VMU}. 44 cases digitized with a 3DHistech Pannoramic Scan II at 40$\times$ magnification (0.25 $\mu m$/px). The \ac{MC} is part of the grading scheme by \cite{mcniel1997evaluation}.
    \item Domain C: Canine lymphoma{, a mesenchymal tumor with round cell morphology}, retrieved from the \ac{DA} of \ac{VMU}. 55 cases digitized with a 3DHistech Pannoramic Scan II at 40$\times$ magnification (0.25 $\mu m$/px). \ac{MC} is part of the grading scheme by \cite{valli2013canine}.
    \item Domain D: Canine cutaneous mast cell tumor, {a mesenchymal tumor with round cell morphology}, retrieved from the \ac{DA} of FUB. 50 cases digitized with an Aperio ScanScope CS2 at 40$\times$ magnification (0.25 $\mu m$/px). \ac{MC} is part of the grading scheme by \cite{kiupel2011proposal}.
    \item Domain E: Human pancreatic and gastrointestinal neuroendocrine tumor{, a tumor with aggregated cell pattern/morphology}, retrieved from the \ac{DA} of \ac{UMC} Utrecht. 55 cases digitized with a Hamamatsu XR (C12000-22) at 40$\times$ magnification (0.23 $\mu m$/px). \ac{MC} is part of the 2022 WHO classification scheme of endocrine and neuroendocrine tumors \citep{who2022endocrine}.
    \item Domain F: Human melanoma, { a neuroectodermal tumor comprising all three morphological patterns, } retrieved from the \ac{DA} of UMC Utrecht. 51 cases digitized with a Hamamatsu XR (C12000-22) at 40$\times$ magnification (0.23 $\mu m$/px). MC is part of the staging and classification scheme of the AJCC for melanoma \citep{amin2017ajcc}. This domain was not labeled and only provided as an additional source of data diversity for unsupervised approaches.
\end{itemize}

While, ideally, a consecutive selection of cases would be desirable to provide representative samples, we intentionally deviated from this norm in this iteration of the challenge. Specifically, we ensured the inclusion of a minimum number of mitotically active cases across all domains. This was done in order to ensure sufficient dataset support for \ac{MF} objects in each respective domain.

We prepared a small {(20 cases) }preliminary test set to check the validity of the algorithmic approaches through the docker submission system. In this dataset, the following domains were included:
\begin{itemize}
    \item Domain $\alpha$: Human breast carcinoma{,  an epithelial tumor with aggregated cell pattern/morphology}, similar to the training set domain A, but scanned with a Hamamatsu RS2 scanner. Five cases, previously used as part of the preliminary test set of \ac{MIDOG} 2021 \citep{aubreville2023mitosis}.
    \item Domain $\beta$: Canine osteosarcoma, {a mesenchymal tumor with predominantly spindle cell morphology,} retrieved from the \ac{DA} of \ac{VMU}. Five cases digitized with a 3DHistech Pannoramic Scan II at 40$\times$ magnification (0.25 $\mu m$/px).
    \item Domain $\gamma$: Human lymphoma{, a mesenchymal tumor, round cell morphology}, retrieved from the \ac{DA} of UMC Utrecht. Five cases digitized with a Hamamatsu XR (C12000-22) at 40$\times$ magnification (0.23 $\mu m$/px).
    \item Domain $\delta$: Canine pheochromocytoma, {a neuroendocrine tumor with aggregated cell pattern/morphology,} retrieved from the \ac{DA} of \ac{VMU}. Five cases digitized with a 3DHistech Pannoramic Scan II at 40$\times$ magnification (0.25 $\mu m$/px).
\end{itemize}

For the evaluation of the challenge, we constructed the so-called final test set, where only a single evaluation per team was permitted. The dataset, { of which an overview is shown in Fig. \ref{fig:overview_domains_testset}, }included 10 cases per domain, encompassing the following domains, evenly divided between human and veterinary samples:

\begin{itemize}
    \item Domain 1: Human melanoma, {  a neuroectodermal tumor comprising all three morphological patterns (round cells, spindle cells, aggregated cell pattern),} retrieved from the \ac{DA} of UMC Utrecht, digitized using a Hamamatsu S360 (C13220) at 40$\times$ magnification (0.23 $\mu m$/px). \ac{MC} is part of the staging and classification scheme of the AJCC for melanoma \citep{balch2009final}.
    \item Domain 2: Human astrocytoma, { a neuroectodermal tumor with round nuclear shape and star-like cytoplasmic projections (mostly fitting into the round cell category) } retrieved from the \ac{DA} of the Institute of Neuropathology at University Hospital Erlangen, digitized with a Hamamatsu S60 at 40$\times$ magnification (0.22 $\mu m$/px). \ac{MC} is part of the 2016 WHO grading scheme \citep{Louis2016}.
    \item Domain 3: Human bladder carcinoma, { an epithelial tumor with aggregated cell pattern,} retrieved from the \ac{DA} of the Institute of Pathology at University Hospital Erlangen, digitized with a 3DHistech Pannoramic Scan II at 40$\times$ magnification (0.25 $\mu m$/px). \ac{MC} is used in the differentiation of tumor types according to \citep{epstein1998world} and was recently confirmed to be prognostically significant by \citep{akkalp2016prognostic}.
    \item Domain 4: Canine breast carcinoma{,  an epithelial tumor with aggregated cell pattern}, retrieved from the \ac{DA} of \ac{VMU}, digitized with a 3DHistech Pannoramic Scan II at 40$\times$ magnification (0.25 $\mu m$/px). \ac{MC} is part of the grading scheme by \cite{pena2013prognostic}.
    \item Domain 5: Canine cutaneous mast cell tumor, {a mesenchymal tumor with round cell morphology, }retrieved from the \ac{DA} of \ac{FUB}, digitized with a Hamamatsu S360 (C13220) at 40$\times$ magnification (0.23 $\mu m$/px). \ac{MC} is part of the grading scheme by \cite{kiupel2011proposal}.
    \item Domain 6: Human meningioma, {a mesenchymal/neuroecodermal tumor with spindle cell shape, } retrieved from the \ac{DA} of the Institute of Neuropathology at University Hospital Erlangen, digitized with the Hamamatsu S60 at 40$\times$ magnification (0.22 $\mu m$/px). \ac{MC} is part of the 2016 WHO grading scheme \citep{Louis2016}. 
    \item Domain 7: Human colon carcinoma,{  an epithelial tumor with aggregated cell pattern,} retrieved from the \ac{DA} of UMC Utrecht,  digitized using a Hamamatsu S360 (C13220) at 40$\times$ magnification (0.23 $\mu m$/px). 
    \ac{MC} is not part of the grading scheme but was shown to predict survival for lymph-node negative colon carcinoma by \cite{sinicrope1999apoptotic}.
    \item Domain 8: Canine splenic hemangiosarcoma, {a mesenchymal tumor with spindle cell morphology,} retrieved from the \ac{DA} of \ac{VMU}, digitized with a 3DHistech Pannoramic Scan II at 40$\times$ magnification (0.25 $\mu m$/px). \ac{MC} is part of the grading scheme of \cite{ogilvie1996surgery}.
    \item Domain 9: Feline (sub)cutaneous soft tissue sarcoma{,  a mesenchymal tumor with spindle cell morphology}, retrieved from the \ac{DA} of \ac{VMU}, digitized with a 3DHistech Pannoramic Scan II at 40$\times$ magnification (0.25 $\mu m$/px). \ac{MC} is part of the grading scheme of \cite{dobromylskyj2021prognostic}.
    \item Domain 10: Feline gastrointestinal lymphoma{, a mesenchymal tumor with round cell morphology}, retrieved from the \ac{DA} of \ac{VMU}, digitized with a 3DHistech Pannoramic Scan II at 40$\times$ magnification (0.25 $\mu m$/px). For cats, the \ac{MC} is know to be correlated with the grade according to the National Cancer Institute working formulation \citep{valli2000histologic}. 
\end{itemize}

While human melanoma (unlabeled) and canine cutaneous mast cell tumor were already part of the training set, the test set used different scanners for both tumor types.

\subsection{Establishment of ground truth}
The \ac{MC} is typically assessed on an \ac{ROI} of 10 high power fields, the size of which is dependent on the optical properties of the microscope \citep{fitzgibbons2023protocol}. For digital microscopy, it is more sensible to directly define the area, calculated from the resolution of the digitization device, which we set in accordance with previous work \citep{veta2019predicting,veta2015assessment} to $2~mm^2$. The \ac{ROI} was selected from each digitized \ac{WSI} by a pathologist with expertise in tumor pathology (C.A.B.) as the area with appropriate tissue and scan quality and the perceived highest mitotic activity, which was considered to be more likely found in a region with high cellular density. This is in accordance with current guidelines \citep{donovan2021mitotic,avallone2021review,ibrahim2022assessment,fitzgibbons2023protocol}. 

Given the well-known inter-rater disagreements in identification and annotation of \acp{MF}, strategic study design methods are essential to limit the effects of these factors on the ground truth for subsequent (ideally unbiased) evaluation. Two main annotation biases need to be considered: When presented with a \ac{MF}, previously identified as such by another expert, an independent expert might be subject to a confirmation bias. Similarly, it has been reported that the chance of overlooking individual \acp{MF}, especially in densely populated cell areas or under sub-optimal image quality, should not be neglected \citep{Bertram2021VetPathol}. 

{  Our annotation method (described in more detail in \cite{bertram2019large}) takes both factors into account by identifying all candidate objects (i.e., \acp{MF}) as well as \acp{NMF}/imposters and then independently rating them by three experts. For the identification, an expert (C.A.B.) initially identified \ac{MF} objects and a roughly similar amount of \ac{NMF} objects in the images. To avoid missing \acp{MF} in this identification step, we trained a RetinaNet \citep{lin2017focal} single-stage object detection model on the annotations of this expert and carried out model inference in a cross-validation scheme to spot additional candidate objects that were previously overlooked. These additional objects were then also assessed by the first expert to yield a first label for each object. At the end of this first step, both classes had a similar prevalence according to the initial assessment, which was not communicated to the experts conducting the consecutive assessments of the \ac{MF}/\ac{NMF} cells. Next, a secondary expert (R.K.) who was blinded to the assessment of the first expert assessed all previously identified objects according to the same two classes (\ac{MF} vs. \ac{NMF}). In case of non-consensus amongst those two experts, a third expert (T.A.D.) was presented the object in question without any information about previously assigned labels to render the final vote.}

All three experts have more than five years of experience in \ac{MF} identification. This independent vote counteracts a confirmation bias, while use of the machine-learning support mitigates the omission of individual objects. Prior to the assessment, the experts agreed on common criteria for the identification of \acp{MF} \citep{donovan2021mitotic}. The annotation of all parts of our dataset (training set, preliminary test set, and the final challenge test set) was carried out using the same methodology. This ground truth definition was used for performance evaluation and ranking of the participants during the \ac{MIDOG} 2022 challenge.

\subsection*{\ac{PHH3}-assisted ground truth}
\begin{figure}
\includegraphics[width=\linewidth]{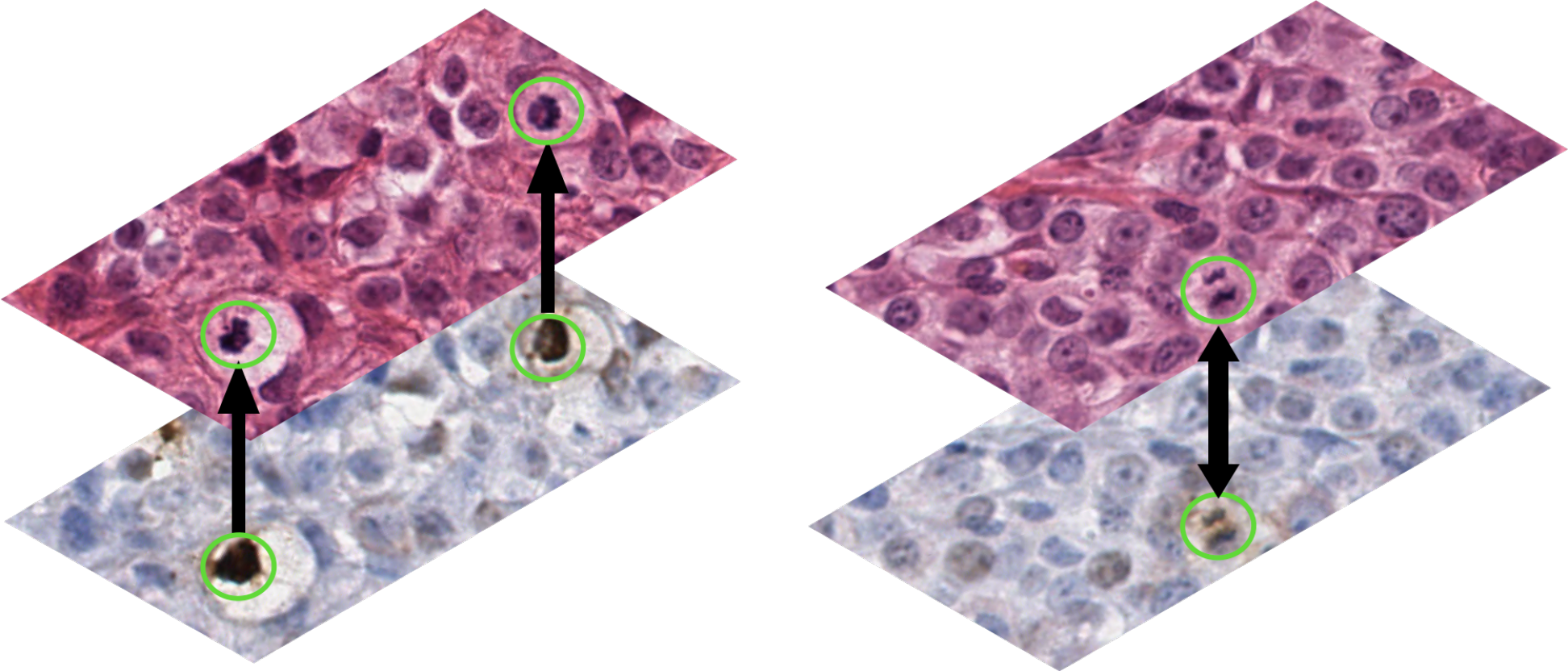}
\caption{Correspondence between hematoxylin and eosin (H\&E)-stained tissue (top) and immunohistochemistry stain against phospho-histone H3 (\ac{PHH3}, bottom). The left panel shows two tumor cells (green circles) with clear immunopositivity against \ac{PHH3}  conclusive for \acp{MF}, supporting \ac{HE} morphology. The right panel shows a mitotic figure in telophase where the \ac{PHH3}-stain is less conclusive, but the morphology in the \ac{HE} is characteristic.}
\label{fig:phh3}
\end{figure}

Due to the known high degree of inter-rater disagreement for mitotic figure assessment \citep{Meyer:2005cl}, it is prudent to create a ground truth that relies less on the subjective judgments of experts. Hence, as an alternative ground truth for the test set, we used \ac{IHC} staining for \ac{PHH3} as a decision support for annotations by a single expert. This ground truth definition was not available during the \ac{MIDOG} 2022 challenge and was developed for this summary paper to gain a better understanding of the algorithmic performance.  
Histone H3 is a protein that is phosphorylated in the early stages of the mitotic phase and represents a specific marker for mitosis \citep{hendzel1997mitosis,bertram2020computerized,tellez2018whole}. However, the specific stain is less pronounced in the last phase (telophase) of mitosis, a phase which is usually morphologically conspicuous with the \ac{HE} stain, and is already present in early prophase, which is usually not apparent based on \ac{HE} morphology. Thus this \ac{IHC} stain cannot be used alone for annotating mitotic figures according to definitions of the \ac{HE} morphology. We hypothesized that the combination of these two staining techniques would increase label consistency. To evaluate \ac{HE} and \ac{PHH3} in the same cells, we de-stained the \ac{HE}-stained slides after digitization and re-stained them with an antibody for \ac{PHH3}, combined with a secondary antibody equipped with a tailored enzyme that reacts with a substrate to yield a brown stain (see Fig. \ref{fig:phh3}). After digitization of the \ac{IHC}-stained slide and subsequent manual registration of both scans, a tool based on the EXACT annotation server was employed by an expert \citep{marzahl2021exact}, in which both scans could be superimposed with variable transparency. Hence, it was possible to simultaneously evaluate both the specific immunopositivity for \ac{PHH3} as well as the morphology in the \ac{HE} stain for each cell. In case of non-perfect registration between cells in the \ac{PHH3} and \ac{HE} stain, the expert annotated the exact coordinate of the \ac{MF} in the \ac{HE} stain. Out of 100 cases of the test set, we were able to register 98 to the respective \acp{ROI} in the \ac{HE} image. For two cases (068 and 100) restaining with \ac{PHH3} was not possible due to damage during tissue handling. Immunopositive cells lacking \ac{HE} morphology of \acp{MF} were not annotated (mostly early prophase \ac{MF}s) as it is impossible to identify them in the \ac{HE} images. { A considerable number of objects was ambiguous from \ac{HE} alone, in which cases  the \ac{IHC} staining pattern was used to decide on these borderline objects.}  

\begin{figure*}
\includegraphics[width=\linewidth]{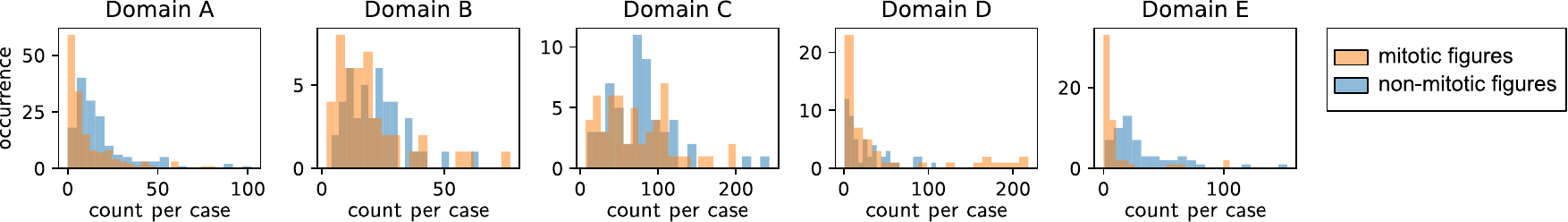}
\caption{Histogram of \acp{MF} and \acp{NMF} in the training set of \ac{MIDOG} 2022.}
\label{fig:histo_training}
\end{figure*}

\subsection{Dataset statistics}

\begin{figure}
    \includegraphics[width=\linewidth]{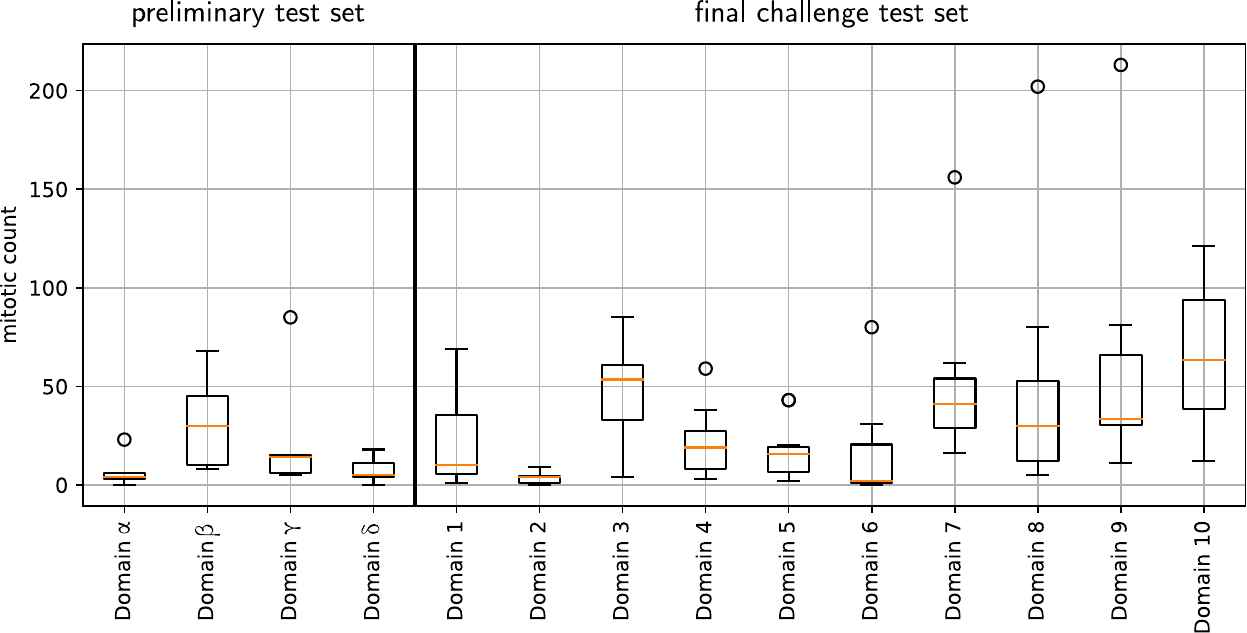}
    \caption{Box-whisker plot of the distribution of \ac{MC} across the domains of the preliminary test set and the final challenge test set. Boxes indicate lower and upper quartile values, colored lines indicate median values.}
    \label{fig:boxplots}
\end{figure}

The \ac{MC} is expected to vary across tumor types and species. This expectation was confirmed in the distribution of \ac{MC} shown in the histogram for the training set (Fig. \ref{fig:histo_training}) as well as in the box-whisker plots for the preliminary and the final challenge test set (Fig. \ref{fig:boxplots}). Tumor types with a comparatively high \ac{MC} in our samples were canine lung cancer (domain B), canine lymphoma (domain C), canine osteosarcoma (domain $\beta$), as well as human bladder carcinoma (domain 3), human colon carcinoma (domain 7), canine hemangiosarcoma (domain 8), and both feline tumors (domains 9 and 10). 
The mean \ac{MC} of the training, preliminary test, and final challenge test set were 26.84, 18.00, and 34.74, respectively. 

\subsection{Reference approaches}

For optimal familiarization, challenge participants were provided with three baseline approaches with algorithmic descriptions and preliminary test results. Out of these three approaches, two were based on the RetinaNet \citep{lin2017focal} single-stage object detection architecture and one was based on the Mask RCNN \citep{he2017mask} architecture. The first RetinaNet-based approach used a domain-adversarial \citep{ganin2016domain} branch and was trained solely on the \ac{MIDOG} 2021 training set (i.e., the identical setting as the reference approach for the \ac{MIDOG} 2021 challenge) and the reference approach for the \ac{MIDOG} 2021 challenge \citep{wilm2022}. Considering that this approach was only trained on human breast cancer, we expected a considerable domain gap. The second RetinaNet-based approach was trained on the six domains of the training set (A-F) and used additional stain augmentation, based on Macenko's method for stain deconvolution \citep{macenko2009method}. As the top-performing approaches of \ac{MIDOG} 2021 were all using (instance) segmentation, we also included the Mask RCNN for this purpose. This approach was, however, not trained with any specific domain-generalizing methods besides default image augmentation. We provided a detailed description of both approaches as part of the challenge proceedings \citep{ammeling2022reference}. 

\subsection{Evaluation methods and metrics}

\ac{MF} identification is a balanced pattern recognition problem in that both an over- and an underestimation of the \acp{MC} can lead to equally detrimental consequences: overestimation may lead to excessively aggressive treatment with significant side effects, whereas an underestimation may contribute to more conservative treatment, potentially diminishing the overall treatment outcome. As in prior challenges \citep{veta2019predicting,veta2016mitosis,LUDOVIC20138,roux2014mitos}, we thus decided to use the $F_1$ score as our primary metric, as it represents the geometric mean between precision and recall and thus benefits from a good operating point set as a balance between both. To counter averaging effects from the strongly heterogeneous distribution of the \ac{MC}, we opted to calculate the $F_1$ score across all cases/images from the summary of respective true positives, false positives, and false negatives over all slides. As the $F_1$ score is { calculated from thresholded results}, it is additionally insightful to see if competing approaches only chose an unsuitable decision threshold while having an otherwise proper pattern discrimination. Hence, we additionally evaluated the \ac{AP} metric, calculated as the mean precision for 101 linearly spaced recall values between 0 and 1. 
Further, we calculated the precision and recall for all algorithms.

{
 
\subsection{Statistical analysis of the results}

To compare the performance of the approaches, an omnibus test using analysis of variance (ANOVA) was first conducted, as is standard procedure. Given significant differences, the test was followed by a Tukey's honest significant difference (HSD) test \citep{tukey1949comparing} as post-hoc test for a pairwise comparison. 
Both tests were performed on the respective $F_1$ score per individual image of the test set. As significance level, we chose $\alpha = 0.05$, as commonly done. The tests allowed us to determine if there were statistically significant differences between the approaches in terms of their $F_1$ scores. In contrast to multiple pairwise t-tests, Tukey's method is inherently controlling for the family-wise error rate. 

Additionally, we performed a mixed linear model regression analysis to analyze effects that can be attributed to the tumor domain or groups therein. For the purpose of our study, we categorized tumors into distinct groups based on tumor morphology, species, and scanner used for digitization. Each of those categories had a previously unseen condition (e.g., an unknown species, scanner, or morphology) in the test set. Morphology was differentiated between aggregated cell pattern, round cell shapes, and spindle cell morphology. 
We differentiated species between human, canine, and feline. Scanners were differentiated between the 3DHistech scanner and the Hamamatsu S360 and S60 scanners. Within this framework, the $F_1$ score was selected as the dependent (endogenous) variable to assess the predictive success of each model. Since each teams' algorithmic contribution can be thought of as a random draw from a larger population of algorithms that are not separate and independent, we accounted for the variance attributed to the different teams by incorporating it as a random effect for the intercept, resulting in a linear mixed effects model with a random intercept and fixed slope. This means that each team gets its own intercept estimate but has a common slope. The models were fitted using the restricted maximum likelihood method and we report the residual variance and \ac{BIC} values for each fit.
}

{ To investigate the distribution across samples, tumor domains and methods, we performed empirical} bootstrapping of the results of each test case, i.e., we randomly selected the same number of cases with replacement from the set of results per case before calculating the precision, recall, \ac{AP} and $F_1$ values. 
{
Bootstrapping offers a robust alternative to individual image-based metrics for statistical analysis, particularly when evaluating metrics like the \ac{AP} and $F_1$ scores in contexts where the target class, such as \acp{MF}, varies widely in prevalence. 

The use of bootstrapping mitigates the disproportionate influence that the target class frequency within individual images might exert on our results, as the $F_1$/AP scores are now calculated on instances comprising multiple images. 
By this methodological choice, we thus obtain values that are empirically aligned with the expected distributional characteristics of the dataset, while simultaneously reducing our dependence on the variable prevalence of mitotic figures in individual images. 

From the bootstrapped (across tumor domain and team) sets, we additionally determined the 80\,\% confidence levels of precision and recall for each team and tumor. This was performed by using a Gaussian kernel density estimator over the bootstrapped sample of precision and recall values and thresholding at an interval to include 80\,\% of the respective values. We chose 80\% as interval, since it facilitated a more easy comparison of the approaches than using larger intervals.

}


\section{Overview of submitted methods}
15 registered users from twelve teams submitted at least once to the preliminary test phase of the challenge. Out of those, nine also submitted to the final test phase.
All models were based on methods of deep learning. The submitted methods were, as in previous challenges, discrepant in more than one key factor, which makes a direct identification of components for a successful \ac{MF} detection method difficult. All teams submitted to track 1 (without additional data) of the challenge, while two teams opted to also submit approaches trained by utilizing additional data. In the following section, we will compare the algorithmic strategies of all teams and subsequently discuss the datasets that were additionally used in track 2. 

\begin{table*}
\resizebox{\textwidth}{!}{%
\begin{tabular}{p{3.5cm}lp{2.5cm}p{4cm}p{2.5cm}cccccp{2.5cm}p{2cm}}
team  & tracks & 1st stage & architecture & second stage & ensembling & TTA & \multicolumn{4}{c}{augmentation} & use of unlabeled domain \\
& & & & & & & geometric & stain & color & other & \\
\hline 
Baseline 1 \citep{ammeling2022reference} & 2  & instance segmentation & Mask RCNN \citep{he2017mask}, ResNet50 backbone & -- & \xmark & \xmark & \cmark & \xmark & \cmark & \xmark & --\\
Baseline 2 \citep{ammeling2022reference} & 1 &  object detection & RetinaNet \citep{lin2017focal}, ResNet18 backbone & -- & \xmark & \xmark & \cmark & \cmark & \cmark & \xmark & domain-adversarial\\
Baseline MIDOG21 \citep{wilm2022} & 1   & object detection & RetinaNet \citep{lin2017focal}, ResNet18 backbone & -- & \xmark & \xmark & \cmark &  \xmark & \cmark & \xmark & domain-adversarial\\
\hline
TIA Centre \citep{jahanifar2022stain}& 1, 2  & segmentation & Efficient-UNet (B0) \citep{jahanifar2021robust} &  EfficientNet-B7 & \cmark &  \cmark & \cmark & \cmark & \cmark  & sharpness  & --\\
TCS Research (RnI) \citep{kotte2022deep}  & 1 & object detection & DETR \citep{carion2020end}, ResNet50-DC5 backbone &  EfficientNet-B7 & \cmark & \xmark & \cmark & \xmark & \cmark & \xmark & -- \\  
USZ / UZH Zurich (ML) \citep{lafarge2022fine} & 1 & classification (sliding window)& P4-ResNet70 \citep{cohen2016group,he2016deep} & -- & \cmark & \xmark & \cmark & \xmark & \cmark & \xmark & hard-negative mining \\
UCLA-HCI \citep{gu2022detecting} & 1 & classification (large tiles) & EfficientNet-B3 & weakly supervised localization & \xmark & \xmark & \cmark & \cmark & \cmark & blur, noise, balanced-mixup & treated as negative\\ 
SKJP \citep{kondo2022tackling} & 1 & object detection & EfficientDet \citep{tan2020efficientdet}, EfficientNet V2-L backbone & -- & \xmark & \xmark & \cmark & \xmark & \xmark & stain normalization & -- \\
HTW Berlin \citep{annuscheit2022radial} & 1 & object detection & YOLO v5 \citep{glenn_jocher_2022_7347926} & -- & \xmark & \cmark & \cmark & \cmark & \cmark & blur/sharpening & domain generalization \\
AI\_medical \citep{sen_yang_2022_7035741, wang2023generalizable} & 1, 2 & segmentation & SK-UNET \citep{wang2021sk}, SE-ResNeXt50 encoder & -- & \xmark & \xmark & \cmark & \xmark & \cmark & Fourier-domain augmentation & -- \\
Virasoft \citep{bozaba2022} & 1 & object detection & YOLO v5 \citep{glenn_jocher_2022_7347926}, CSPDarknet53 stem & EfficientNet-B3 & \xmark & \xmark & \cmark & \xmark & \xmark & mosaic & -- \\ 
HITszCPATH \citep{wang2022multi} & 1 & object detection & RetinaNet \citep{lin2017focal}, ResNet50 stem &  -- & \xmark & \xmark & \cmark & \xmark & \cmark & -- & auxiliary classifier \\
\end{tabular}
}
\caption{Overview of the submitted methods by all participating teams. TTA indicates test-time augmentation.}
\label{tab:overview}
\end{table*}

\subsection{Pattern recognition tasks}
The majority of teams (5/9) chose to frame the task as an object detection task (see Table~\ref{tab:overview}), partially with a second classification stage. Two teams used a semantic segmentation approach and two teams chose a classification-based detection. In particular, the approach by \cite{jahanifar2022stain} used fixed-size disks around the centroid coordinate of the \acp{MF} to generate the segmentation {  mask target} for track 1 and a segmentation mask generated by the NuClick algorithm \citep{koohbanani2020nuclick} for track 2, while the approach by \cite{sen_yang_2022_7035741} used the filled inner circle of the provided bounding box as segmentation target. In contrast, \cite{lafarge2022fine} used a classification of patches (78$\times$78 px) with a sliding window like in the original works by \cite{cirecsan2013mitosis}. \cite{gu2022detecting} framed object localization as a weakly-supervised learning task derived from class activation maps of medium-sized (240$\times$240 px) patches that were classified as containing a \ac{MF} or not. 

\subsection{Architectures}
The majority of submissions were derivatives of \acp{CNN}, while one team designed their method based on the \ac{DETR} \citep{carion2020end}, which is an object detector derived from the vision transformer class of models and hence uses a \ac{CNN} solely for feature extraction. Amongst the other approaches, EfficientNet \citep{tan2019efficientnet}-derived architectures were frequently used. Variants of EfficientNet were used as second stage in the approaches by \cite{jahanifar2022stain}, \cite{kotte2022deep}, and \cite{bozaba2022}, as classification approach by \cite{gu2022detecting}, and as a stem of the mitosis detector by \cite{jahanifar2022stain} and \cite{kondo2022tackling}. Other researchers chose different well-established network stems, such as ResNet \citep{he2016deep}, SE-ResNeXt \citep{hu2018squeeze}, or CSPDarknet53 \citep{bochkovskiy2020yolov4}.  

\subsection{Ensembling and Test-Time Augmentation}
While both ensembling and \ac{TTA} are strategies well-known to enhance model robustness, they were only employed by a minority of participants (see Table~\ref{tab:overview}). Only the winning approach by \cite{jahanifar2022stain} employed both ensembling and TTA. The runner-up approach by \cite{kotte2022deep} employed a tailored ensembling of model scores of the first and second stages but only in cases where the score of an object in the first stage did not exceed a given threshold. The approach by \cite{lafarge2022fine} ensembled two models trained with different augmentation strategies, and integrated the effect of 90-degree rotation for TTA via the use of a rotation invariant model \citep{cohen2016group}. The approach by \cite{annuscheit2022radial} used four-fold \ac{TTA} using mirroring of the images. 

\subsection{Augmentation}
All participating teams used standard geometric image transformations like rotation, scaling, and elastic deformations. Additionally, the majority of teams opted to use one form of standard color augmentation that aims at manipulating the hue, brightness, and contrast. Additionally, multiple teams opted to use image perturbations such as blurring, sharpening, and noising. \cite{bozaba2022} additionally employed mosaic augmentation. \cite{bochkovskiy2020yolov4}, and \cite{gu2022detecting} additionally used balanced mixup \citep{galdran2021balanced}. Besides those general computer vision augmentation strategies, specific stain augmentation strategies for \ac{HE}-stained images were employed by three teams \citep{jahanifar2022stain,gu2022detecting,annuscheit2022radial}, while the approach by \cite{yang2020fda} augmented images by performing a style-transfer in the frequency-domain.

\subsection{Use of the unlabeled domain}
The unlabeled domain F of the training set (human melanoma) was employed by four teams. \cite{lafarge2022fine} designed a hard-negative mining scheme that additionally employed the unlabeled domain by un-mixing the stains into hematoxylin, eosin, and a residual component, and then extracted objects with high residual components (e.g., stain artifacts), which can be mistaken for \acp{MF}. \cite{gu2022detecting} used the surplus domain to treat all images as negatives and counteracted these noisy labels with a specifically crafted loss function. \cite{annuscheit2022radial} used the domain as an additional domain in a representation learning scheme for domain adaptation. Finally, \cite{wang2022multi} used the additional data in an auxiliary domain classifier in a multi-task learning scheme. 

\subsection{Domain generalization methodologies}
Besides augmentation, several teams employed specific strategies targeted at domain generalization. \cite{annuscheit2022radial} designed a domain adaptation scheme based on metric learning where the distance of each sample to prototypes of all domains was minimized to achieve domain generalization. \citep{wang2022multi} employed multi-task learning with two auxiliary tasks: an overall \ac{MF} classification for the patch and a tumor domain classifier, likely regularizing the model (and hence counteracting domain overfitting). Similarly, \citep{sen_yang_2022_7035741} added a weight perturbation to the loss term, as this was shown to regularize the model and make it more robust to domain shifts \citep{wu2020adversarial}.

\subsection{Addional datasets used in track 2}
In the second track of the challenge, it was permitted to use publicly available datasets. \cite{sen_yang_2022_7035741} used a Hover-Net \citep{graham2019hover} which was trained on other histopathology datasets to generate more accurate segmentation masks. Similarly, the approach by \cite{jahanifar2022stain} created enhanced \ac{MF} segmentation masks by using NuClick \citep{koohbanani2020nuclick} and additionally by incorporating the TUPAC16 \citep{veta2019predicting} dataset to the training dataset.

\section{Results}
\begin{figure*}[t]
\subfloat[$F_1$]{\scalebox{1}{\label{fig:f1_score}\includegraphics[width=0.46\linewidth]{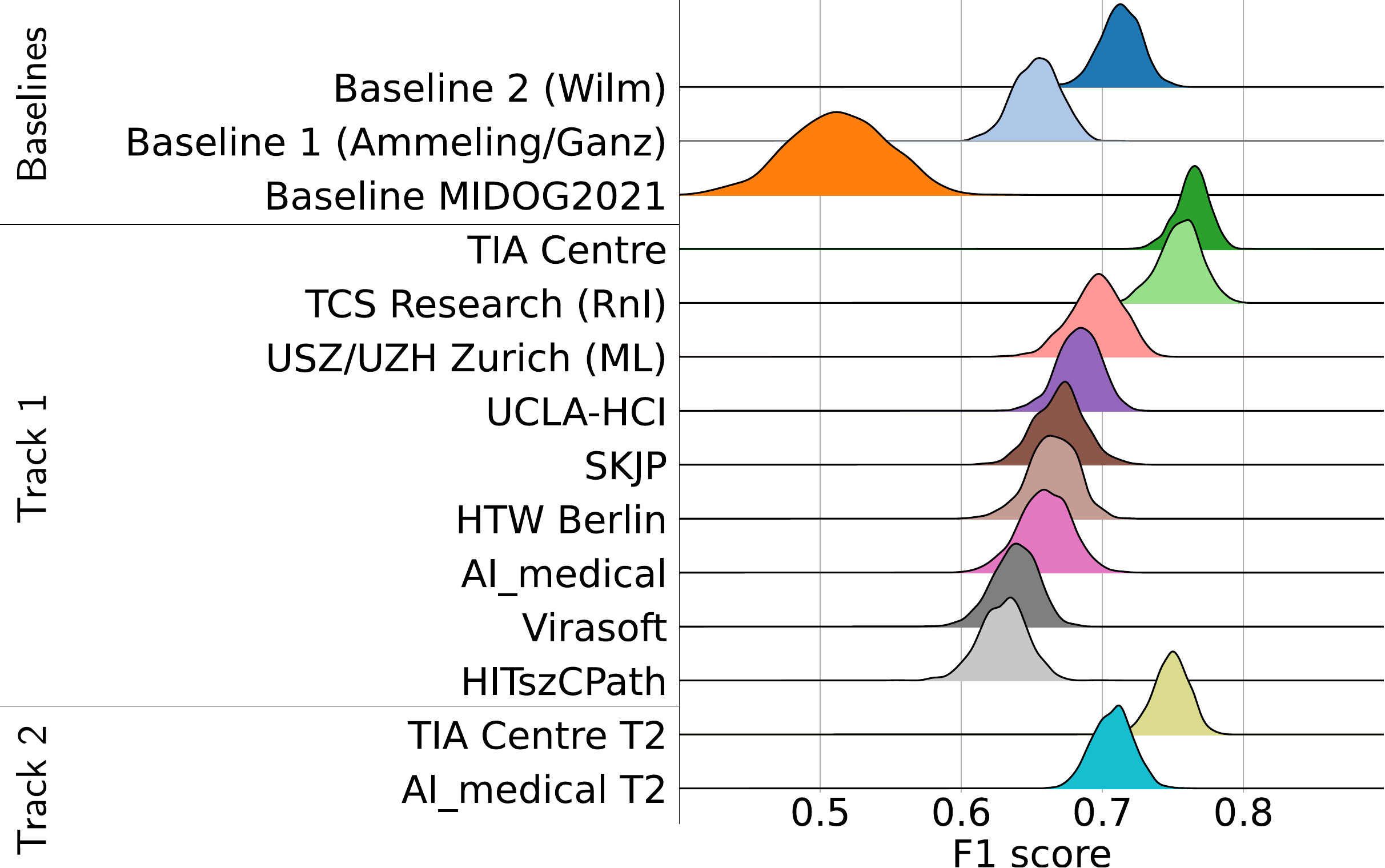}}}\hfill
\subfloat[Average Precision (AP)]{\scalebox{1}
{\label{fig:ap_metrics}\includegraphics[width=0.46\linewidth]{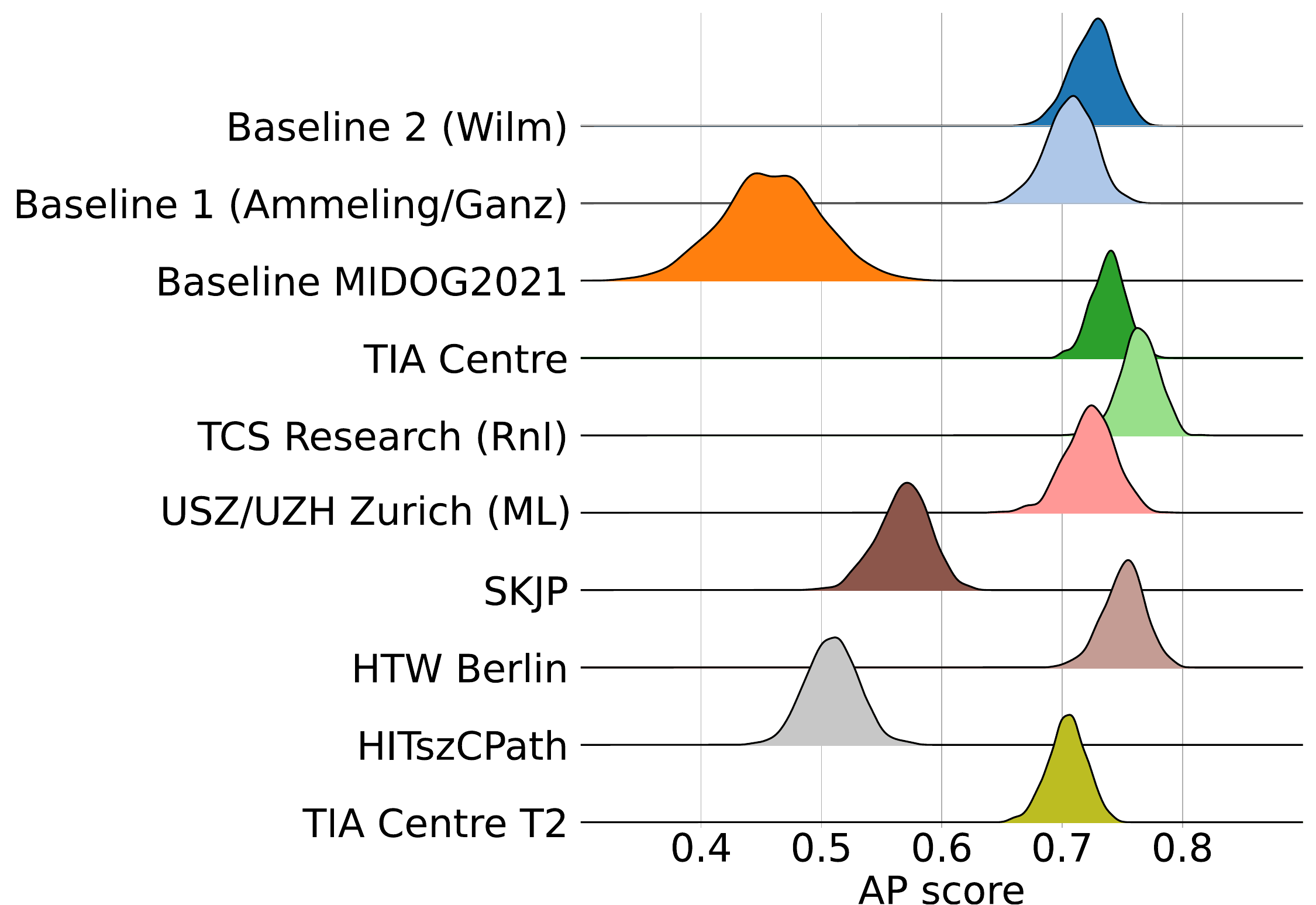}}}
\\
\subfloat[Precision]{\scalebox{1}{\label{fig:precision}\includegraphics[width=0.46\linewidth]{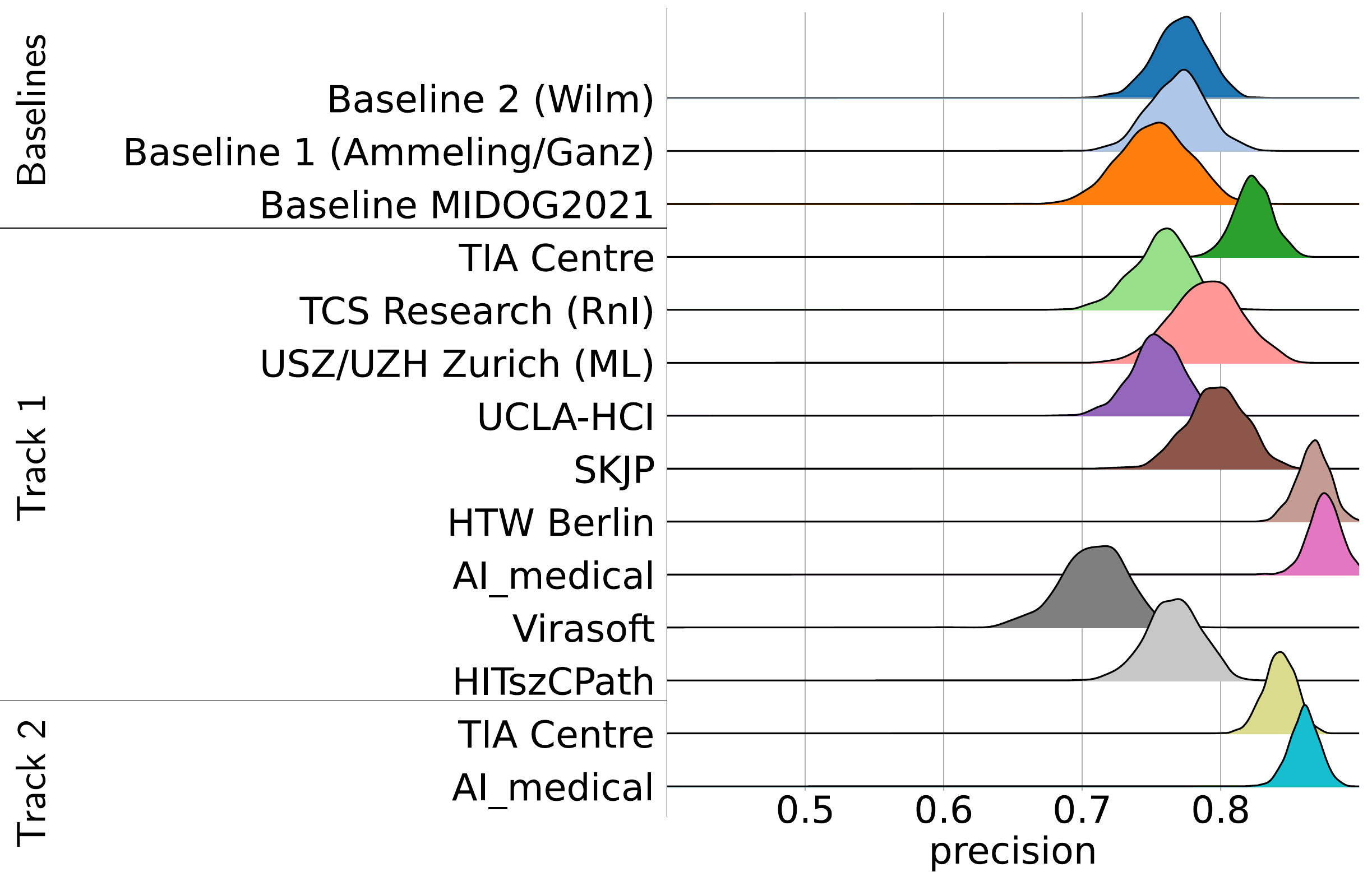}}}\hfill
\subfloat[Recall]{\scalebox{1}{\label{fig:recall}\includegraphics[width=0.46\linewidth]{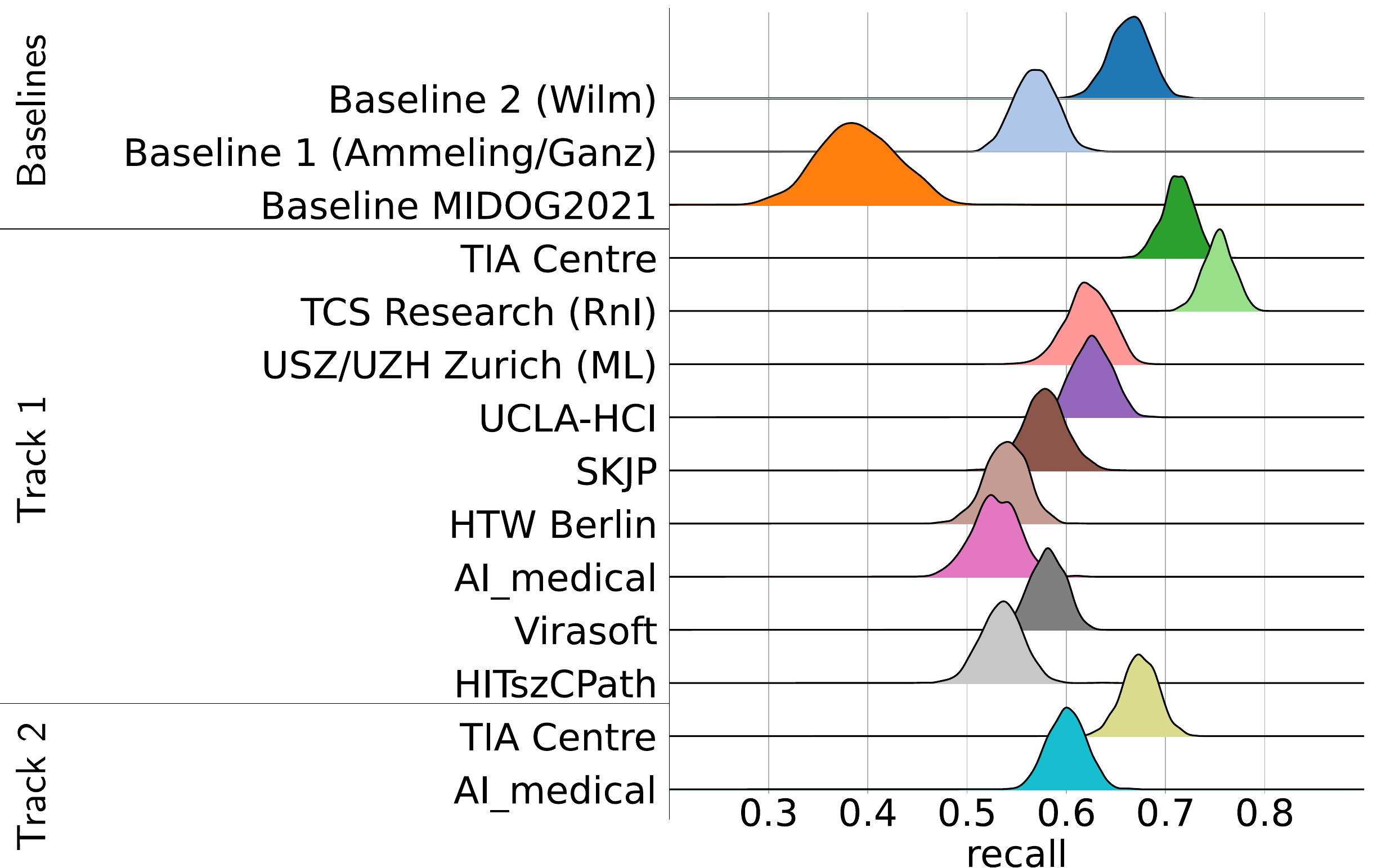}}}
\caption{Distribution of the $F_1$ score, precision, recall, and AP metric as a result of bootstrapping. Only submissions that provided meaningful scores per detection are shown in the AP metric diagram.}

\end{figure*}

\begin{figure}[t]
\includegraphics[width=\linewidth]{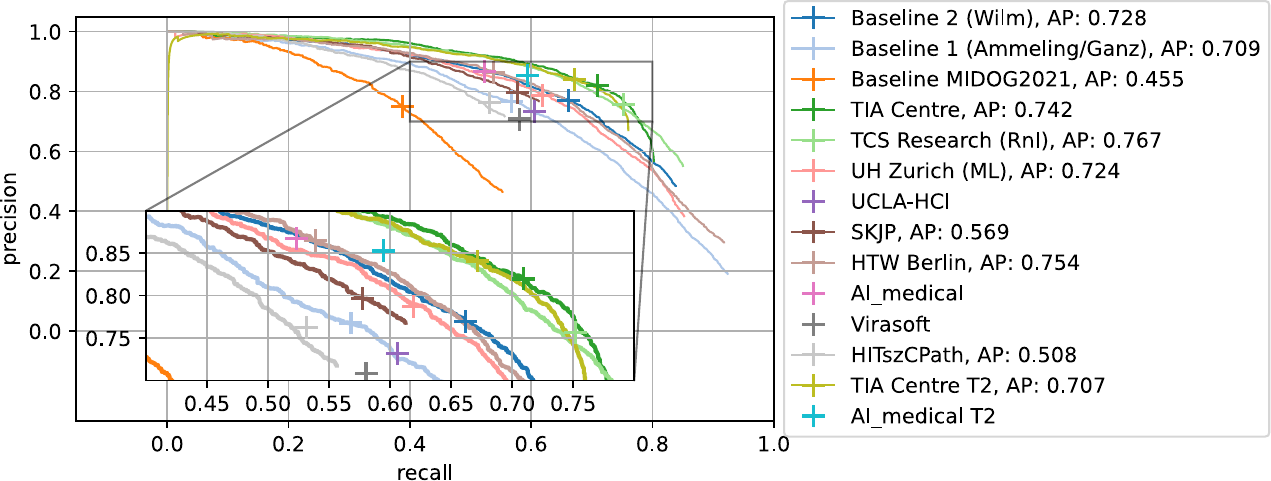}
\caption{Precision-recall values and curves (for all participants where the model score per \acp{MF} was provided and consistent). The marker indicates operating point calculated by the thresholded detections of the participants. Minor mismatches may be explained by post-processing after thresholding.}
\label{fig:pr_curve}
\end{figure}

{  Overall, as shown in Fig. \ref{fig:f1_score} and Fig. \ref{fig:ap_metrics}, we found that two of the approaches submitted to the challenge had outstanding performance. }The evaluation of track 1 on all ten tumor domains of the test set shows that the TIA Center approach \citep{jahanifar2022stain} yielded the best overall performance ($F_1=0.764$), closely followed by the approach from the TCS Research team \citep{kotte2022deep} ($F_1=0.757$). Breaking this down into the ten tumor domains, we find a similar overall picture, with both approaches scoring first or second in all domains (see Table~\ref{tab:F1_across_domains}). We also note that the two leading approaches chose different strategies when optimizing the operating point: While the TIA Center approach yielded a moderately lower recall value at a higher precision value, we found the opposite to be true for the TCS Research approach (see Fig. \ref{fig:precision} and Fig. \ref{fig:recall}). The $F_1$ score is roughly reflected in the precision-recall curves of Fig. \ref{fig:pr_curve}. 

In the second track of the challenge, we find a clear superiority of the approach by \citep{jahanifar2022stain}, further supported by having the leading edge in all tumor domains. 

Comparing the performance in both tracks across tumor domains, we find that tumor domain 2 (human astrocytoma) and 6 (human meningioma), i.e, the neuropathological domains, seemed to have been particularly challenging, with overall maximum $F_1$ scores of 0.63 and 0.68, respectively (see Table~\ref{tab:F1_across_domains}).  On the contrary, the domains 1 (human melanoma), 3 (human bladder carcinoma), 5 (canine cutaneous mast cell tumor), and 8 (canine splenic hemangiosarcoma) were the tumor domains to which the algorithms generalized best, achieving $F_1$ scores of up to 0.82, 0.81, 0.82 and 0.82, respectively.

\begin{table*}
\resizebox{\textwidth}{!}{%
\begin{tabular}{lccccccccccc}
Team & overall &   Tumor 1 & Tumor 2 & Tumor 3 & Tumor 4 & Tumor 5 & Tumor 6 & Tumor 7 & Tumor 8 & Tumor 9 & Tumor 10  \\ 
Baseline 2 (Wilm) & 0.714 [0.68,0.74]&0.74 [0.61,0.79]&0.48 [0.27,0.63]&0.75 [0.69,0.79]&0.68 [0.61,0.73]&0.81 [0.76,0.84]&0.66 [0.50,0.73]&0.72 [0.62,0.78]&0.77 [0.64,0.82]&0.69 [0.55,0.75]&0.66 [0.56,0.72]\\
Baseline 1 (Ammeling/Ganz) & 0.654 [0.62,0.68]&0.72 [0.59,0.78]&0.32 [0.13,0.48]&0.72 [0.66,0.76]&0.56 [0.48,0.61]&0.76 [0.67,0.80]&0.60 [0.42,0.72]&0.67 [0.59,0.72]&0.70 [0.59,0.74]&0.57 [0.40,0.66]&0.64 [0.55,0.72]\\
Baseline MIDOG2021 & 0.513 [0.44,0.58]&0.73 [0.58,0.79]&0.38 [0.13,0.67]&0.74 [0.68,0.79]&0.23 [0.12,0.32]&0.69 [0.64,0.73]&0.62 [0.38,0.70]&0.69 [0.59,0.74]&0.50 [0.34,0.58]&0.32 [0.22,0.36]&0.08 [0.03,0.13]\\
\hline
TIA Centre & \textbf{0.764} [0.74,0.78]& \textbf{0.80} [0.74,0.84]&\textbf{0.65} [0.38,0.79]& \textbf{0.81} [0.78,0.83]& 0.71 [0.62,0.78]&\textbf{0.83} [0.81,0.86]& \textbf{0.71} [0.52,0.78]&\textbf{0.75} [0.65,0.82]& 0.79 [0.70,0.83]&0.70 [0.58,0.77]&\textbf{0.77} [0.70,0.81]\\
TCS Research (RnI) & 0.757 [0.72,0.78]&0.76 [0.66,0.80]&0.48 [0.24,0.72]&0.79 [0.73,0.84]&\textbf{0.73} [0.66,0.76]&0.79 [0.74,0.83]&0.65 [0.41,0.75]&0.74 [0.64,0.80]&\textbf{0.84} [0.73,0.87]&\textbf{0.72} [0.62,0.77]&\textbf{0.77} [0.70,0.82]\\
USZ / UZH Zurich (ML) & 0.696 [0.66,0.73]&0.76 [0.58,0.83]&0.28 [0.10,0.52]&0.75 [0.69,0.80]&0.66 [0.55,0.73]&0.73 [0.66,0.79]&0.64 [0.45,0.71]&0.71 [0.63,0.77]&0.78 [0.66,0.82]&0.64 [0.46,0.73]&0.66 [0.58,0.73]\\
UCLA-HCI & 0.685 [0.65,0.71]&0.55 [0.39,0.67]&0.48 [0.20,0.74]&0.76 [0.70,0.80]&0.68 [0.62,0.72]&0.77 [0.74,0.79]&0.57 [0.46,0.65]&0.69 [0.61,0.75]&0.72 [0.62,0.76]&0.58 [0.44,0.66]&0.73 [0.67,0.79]\\
SKJP & 0.671 [0.64,0.70]&0.76 [0.65,0.80]&0.51 [0.23,0.69]&0.75 [0.69,0.80]&0.60 [0.53,0.65]&0.70 [0.59,0.76]&0.65 [0.48,0.75]&0.67 [0.57,0.73]&0.71 [0.58,0.77]&0.60 [0.52,0.64]&0.63 [0.52,0.71]\\
HTW Berlin & 0.666 [0.63,0.70]&0.75 [0.64,0.80]&0.57 [0.30,0.76]&0.72 [0.64,0.78]&0.57 [0.41,0.66]&0.77 [0.68,0.81]&0.66 [0.51,0.75]&0.64 [0.58,0.69]&0.69 [0.52,0.76]&0.56 [0.40,0.63]&0.68 [0.57,0.75]\\
AI\_medical & 0.659 [0.62,0.69]&\textbf{0.80} [0.70,0.84]&0.63 [0.33,0.83]&0.74 [0.67,0.80]&0.64 [0.52,0.73]&0.79 [0.72,0.84]&0.65 [0.49,0.72]&0.68 [0.62,0.72]&0.68 [0.56,0.75]&0.52 [0.43,0.60]&0.59 [0.51,0.65]\\
Virasoft & 0.639 [0.61,0.67]&0.66 [0.50,0.74]&0.34 [0.15,0.65]&0.70 [0.66,0.72]&0.62 [0.54,0.67]&0.60 [0.54,0.65]&0.59 [0.35,0.65]&0.60 [0.49,0.67]&0.73 [0.64,0.77]&0.63 [0.50,0.68]&0.61 [0.54,0.66]\\
HITszCPath & 0.630 [0.59,0.66]&0.75 [0.65,0.80]&0.38 [0.17,0.60]&0.73 [0.66,0.77]&0.55 [0.44,0.63]&0.72 [0.66,0.76]&0.57 [0.34,0.66]&0.62 [0.56,0.65]&0.70 [0.59,0.76]&0.49 [0.33,0.56]&0.61 [0.53,0.67]\\
\hline
TIA Centre (Task 2) & \textbf{0.749} [0.72,0.77]&\textbf{0.83} [0.77,0.86]& \textbf{0.70} [0.37,0.86]& \textbf{0.81} [0.78,0.84]& \textbf{0.71} [0.61,0.76]& \textbf{0.82} [0.80,0.84]&\textbf{0.69} [0.57,0.74]& \textbf{0.73} [0.64,0.79]& \textbf{0.78} [0.68,0.82]&\textbf{0.67} [0.52,0.73]& \textbf{0.73} [0.65,0.78]\\
AI\_medical (Task 2) & 0.708 [0.68,0.73]&0.82 [0.74,0.86]&0.61 [0.29,0.78]&0.78 [0.71,0.83]&0.65 [0.56,0.71]&0.78 [0.74,0.81]&0.66 [0.50,0.76]&0.71 [0.64,0.75]&0.73 [0.61,0.80]&0.58 [0.42,0.64]&0.71 [0.66,0.77]\\
\end{tabular}
}
\caption{$F_1$ values across all tumor domains for all participants. Values in brackets indicate 95\% confidence interval as a result of bootstrapping. The top group are the baselines, the middle group are the submissions in track 1 and the bottom group are the submissions in track 2 of the challenge.}
\label{tab:F1_across_domains}

\end{table*}

\begin{figure*}
\includegraphics[width=\textwidth]{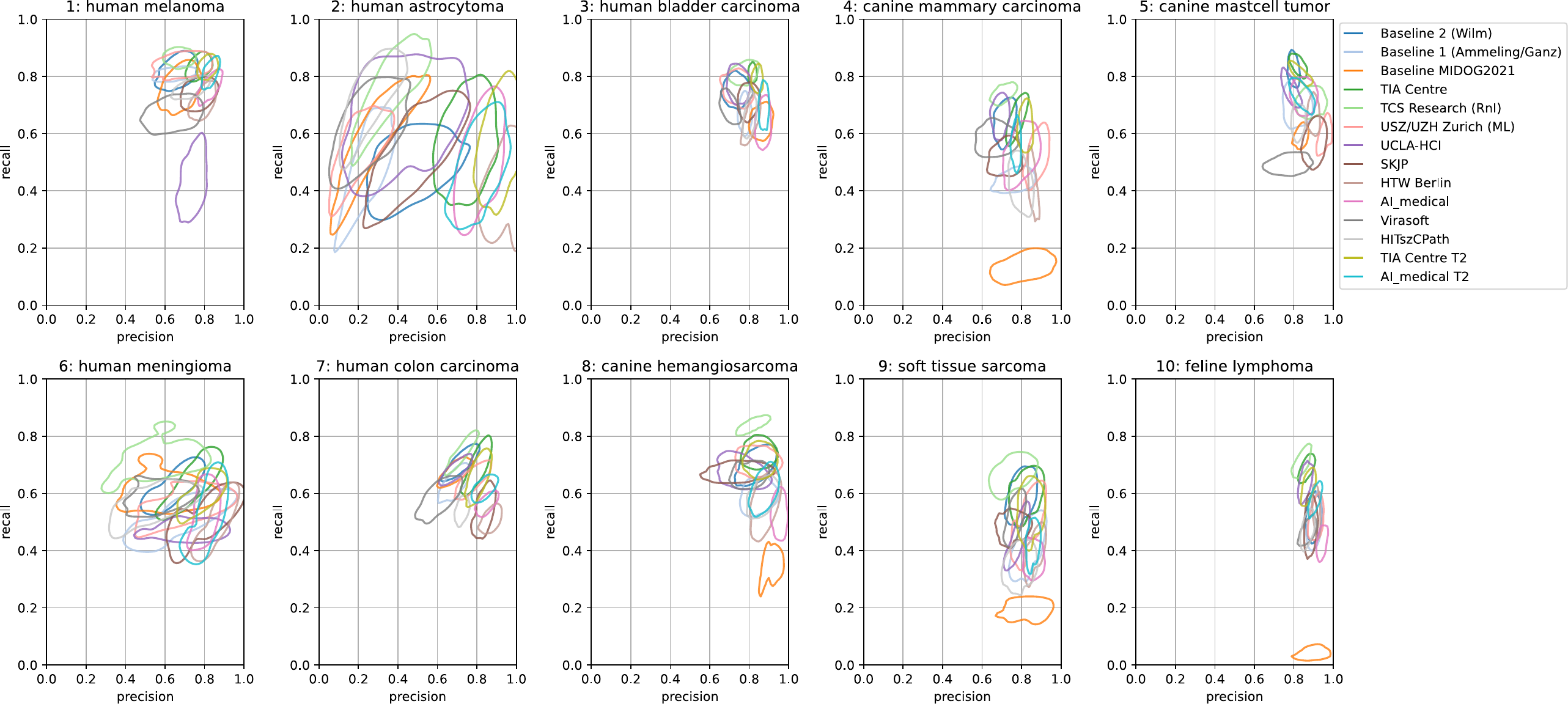}
\caption{80\% confidence regions for precision and recall for each team/approach and tumor type. Confidence intervals were established using a Gaussian kernel density estimator applied to the bootstrapped datasets. Kindly refer to the supplementary material for an alternative version of this Figure, where the subplots compile data over tumor domains for each team. }
\label{fig:confidence_regions}
\end{figure*}

\begin{figure}
\includegraphics[width=\linewidth]{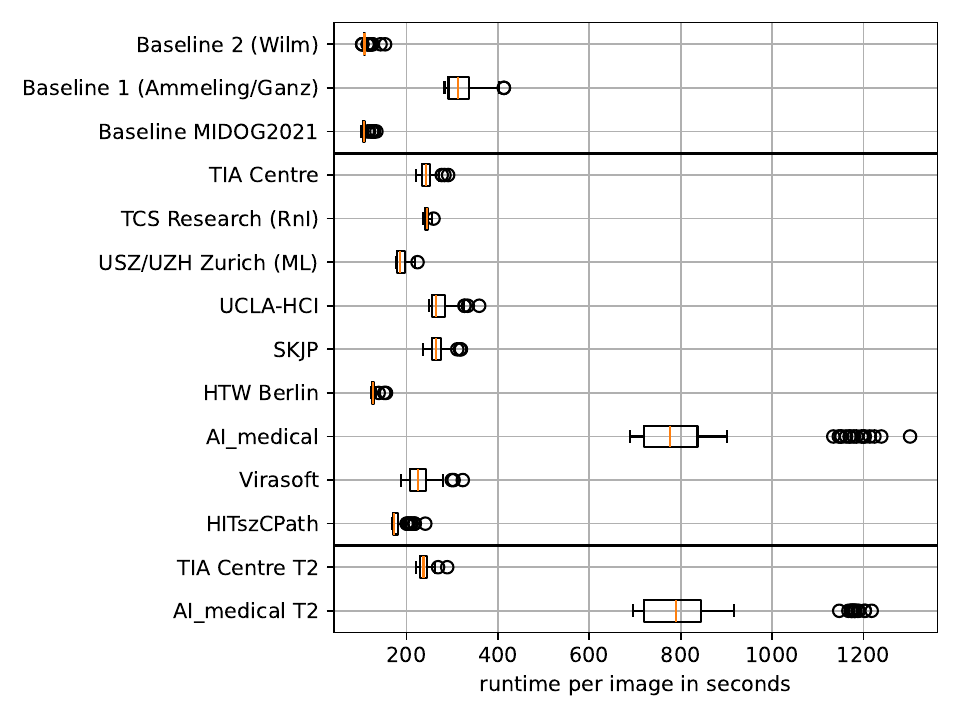}
\caption{ Runtime assessment (box-whisker-plot) for all algorithmic approaches on grand-challenge.org. Boxes represent 25th and 75th percentile, orange line indicates median.}
\label{fig:runtime}
\end{figure}

{ 
\subsection{Statistical analysis}
The ANOVA yielded an F-value of 7.295 ($\mathrm{df}_\mathrm{group}=13$, $\mathrm{df}_\mathrm{residual}=1386$, $p < 0.0001$), indicating a significant difference between at least two of the algorithms.  While the $F_1$ score, which we assessed statistically,  differed considerably between the algorithms, the post-hoc analysis, performed as Tukey HSD hypothesis test (depicted in Fig.~A.2 of the supplementary material), yielded statistically significant ($p<0.05$) differences only between the results of the MIDOG 2021 domain-adversarial RetinaNet baseline and most other approaches (with the exception of the Mask RCNN baseline and the Virasoft / HITszCPath approaches), between the HTW Berlin and the leading TIA Centre  method, and between both the Virasoft and HITszCPath and all approaches by TIA Centre and the approach by TCS Research. While the TIA Centre approaches and the TCS Research approach reached a higher overall F1 score than the MIDOG 2022 domain-adversarial baseline provided by the organizers, this difference was not significant, as of this statistical test.


Visual analysis of the 80\% confidence regions of precision and recall per tumor domain, shown in Fig. \ref{fig:confidence_regions}, reveals a wide spread in performance in the human astrocytoma and the human meningioma domains, both originating from a lab that did not provide samples to the training set. Furthermore, the analysis yields low recall for the baseline MIDOG 2021 RetinaNet approach in feline soft tissue sarcoma, canine hemangiosarcoma, feline lymphoma, and for canine mammary carcinoma. 
The figure also reveals that the UCLA-HCI approach, while performing well overall, had a particularly low recall in the human melanoma case. 

The results of the linear mixed effects model shown in Table~\ref{tab:tumordomain} confirm the visual impression of considerable differences between tumor domains. Since all variables in this regression model are binary projections of the categorical variable and hence mutually exclusive, the coefficients can be interpreted directly as differences in the $F_1$ score. Feline soft tissue sarcoma, canine mammary carcinoma, and human astrocytoma all showed a highly significant decrease in detection performance.
The subgroup analysis of morphological patterns (Table~\ref{tab:tumorgroup}) showed a significantly reduced performance for the group of spindle cell tumors (not seen in training). Similarly, also for the analysis of domains according to the species (Table~\ref{tab:species}), we find a small but significant drop in performance for feline specimens, which were also not part of the training set. Notably, we also found significant differences between the scanners, with the images scanned on the Hamamatsu S360 scanner (which was already seen as part of the training set) performing slightly better than on the Hamamatsu S60 (Table~\ref{tab:scanner}). The variance of the random intercept is small with $0.004$ in all models suggesting that the teams have similar intercepts and that the random intercept contributes little to the overall variability.

\subsection{Runtime analysis}
All inference runs were conducted on the same platform, grand-challenge.org, where all jobs are run on the same environment (ml.g4dn.xlarge configuration, 4 virtual CPU cores, 16 GiB RAM, NVIDIA Tesla T4 with 16GB of VRAM, hosted by Amazon Web Services). This enabled us --- within certain limits --- to also compare the runtime of the approaches. The analysis, shown in Fig. \ref{fig:runtime}, reveals that only three approaches (the HTW Berlin and both RetinaNet approaches) had a runtime in the area of $100\,s$ per image. The approach by USZ/UZH Zurich and HITszCPath reached medium inference time per image of below $200\,s$, while most approaches were in the $200-400\,s$ range. In contrast, the two-stage approach by the AI\_medical team required a considerably higher run-time in the range of $800\,s$ per image.
}

\subsection{Assessment on alternative (\ac{PHH3}-assisted) ground truth}
After the full annotation of 98 cases based on the joint information of the \ac{HE} and \ac{PHH3}-stained images, we found an increase in the count of \ac{MF} by 15.0\%.
Out of the mitotic figures identified aided by the PHH3-stained images, 28.78\% were previously not part of the { majority vote} of the three experts based on the \ac{HE} stain. We performed a post-hoc analysis of all these cells, the results of which are depicted in Table~\ref{tab:phh3_breakdown}. Out of those \ac{MF} only identified with help of the \ac{PHH3} stain, 20.97\% were from the 9\% of cases of feline lymphoma, which are generally difficult due to the small cell size resulting at low cellular details at the given image resolution. Over the complete test set, the primary reason for the discrepancy was a borderline mitotic figure morphology, which was hard to discriminate against imposters due to cells being out of focus or superimposed to other cells in thick tissue sections, poor tissue/image quality such as overstained chromatin structures, prophase morphology without obvious chromatin spikes that are difficult to differentiate from apoptotic cells or other, not further classified reasons.  
Less common were difficulties distinguishing the \ac{MF} from imposters due to a borderline cell cycle phase to the G2-phase with early membrane changes and G1-phase with formation of nuclear membranes of the two neighboring daughter cells. In $5.85\%$ of cases the \acp{MF} were found to have an unusual morphology, while in $0.54\%$ of cases we found an incomplete capture of the cell at the image borders. Only $1.08\%$ of mitotic figures were considered labeling errors in the HE approach, as characteristic \ac{MF} morphology was apparent. 

\begin{figure*}
\includegraphics[width=\linewidth]{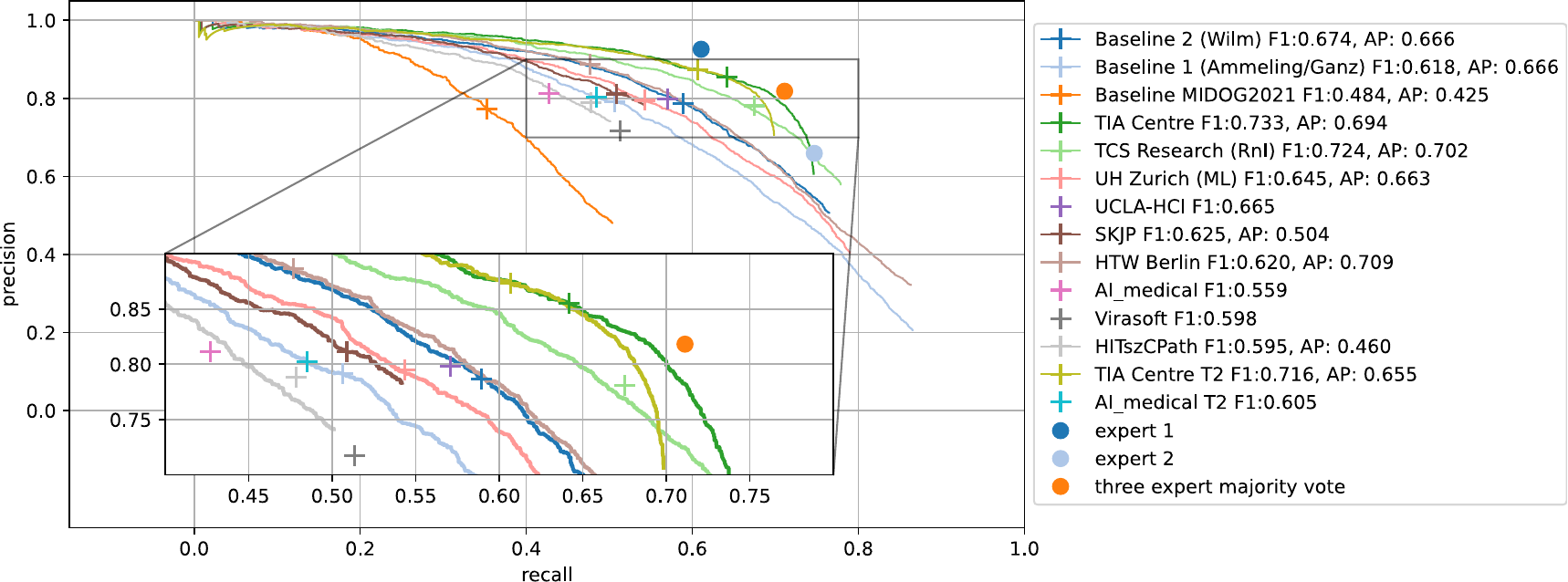}
\caption{Precision and recall of all approaches and the experts, evaluated on the \ac{PHH3}-assisted alternative ground truth. AP values and curves are only given for approaches where the model scores were provided and consistent. The expert scores represent the independent assessment of experts 1 and 2 on the hematoxylin and eosin-stained images, which was performed when establishing the original challenge ground truth, the { majority vote} represents the challenge ground truth.}
\label{fig:phh3-gt}
\end{figure*}

When evaluating with this alternative, \ac{IHC}-assisted ground truth, we found overall lower recall values for all approaches, as shown in Fig. \ref{fig:phh3-gt}, also resulting in overall lower \ac{AP} and $F_1$ values. However, the order of the approaches, when sorted by the $F_1$ value, was almost unaltered. Fig. \ref{fig:phh3-gt} also shows the precision and recall values of both experts using the original \ac{HE} images when evaluated on the \ac{PHH3}-assisted alternative ground truth, as well as the respective values for the three-expert majority vote, indicating a good alignment between the \ac{HE}{-based} and the \ac{IHC}-assisted GT. For expert 1 (C.A.B.), we found an overall precision, recall, and $F_1$ value of 0.926, 0.611, and 0.736, respectively, and for expert 2 (R.K.) we found an overall precision, recall, and $F_1$ value of 0.659, 0.747 and 0.700, respectively. The three expert { majority vote} achieved a precision, recall, and $F_1$ value of 0.818, 0.711, and 0.761 respectively.

\begin{table*}
\begin{center}
\begin{tabular}{lrrrrrr}
\hline
                           &  coef. & std.err. &        z & P$> |$z$|$ & [0.025 & 0.975]  \\
\hline
intercept                  &  0.639 &    0.025 & 25.114 &       0.000 &  0.589 &  0.689  \\
1: human melanoma    &  0.008 &    0.026 &  0.289 &       0.773 & -0.044 &  0.059  \\
2: human astrocytoma & -0.114 &    0.026 & -4.317 &       0.000 & -0.165 & -0.062  \\
3: human bladder carcinoma  &  0.071 &    0.026 &  2.678 &       0.007 &  0.019 &  0.122  \\
4: canine mammary carcinoma      & -0.111 &    0.026 & -4.224 &       0.000 & -0.163 & -0.060  \\
5: canine cutaneous mast cell tumor     &  0.069 &    0.026 &  2.610 &       0.009 &  0.017 &  0.120  \\
6: human meningioma & -0.059 &    0.026 & -2.244 &       0.025 & -0.111 & -0.007  \\
7: human colon carcinoma    &  0.004 &    0.026 &  0.148 &       0.882 & -0.048 &  0.056  \\
9: feline soft tissue sarcoma & -0.135 &    0.026 & -5.122 &       0.000 & -0.187 & -0.083  \\
10: feline lymphoma            & -0.020 &    0.026 & -0.762 &       0.446 & -0.072 &  0.032  \\
group var                  &  0.004 &    0.008 &        &             &        &         \\
\hline
\end{tabular}
\end{center}
\caption{Results from the linear mixed effects model assessing the influence of tumor domain on $F_1$-Score with 'team' as a random effect for the intercept and canine hemangiosarcoma acting as baseline category. The fit of the model was determined using the Restricted Maximum Likelihood (REML) method, with a residual variance of $0.0486$ and a \ac{BIC} of $-100.79$. }
\label{tab:tumordomain}
\end{table*}

\begin{table}
\begin{center}
\resizebox{\linewidth}{!}{%
\begin{tabular}{lrrrrrr}
\hline
                             &  Coef. & Std.Err. &       z & P$> |$z$|$ & [0.025 & 0.975]  \\
\hline
intercept (30 cases)                   &  0.627 &    0.021 & 30.477 &       0.000 &  0.586 &  0.667  \\
all three (10 cases)  &  0.020 &    0.022 &  0.888 &       0.374 & -0.024 &  0.064  \\
round cell (30 cases)    & -0.009 &    0.016 & -0.593 &       0.553 & -0.040 &  0.022  \\
spindle cell (30 cases)  & -0.052 &    0.016 & -3.308 &       0.001 & -0.083 & -0.021  \\
group var                    &  0.004 &    0.008 &        &             &        &         \\
\hline   &             &        &         \\
\hline
\end{tabular}}
\caption{Results from the linear mixed effects model assessing the influence of tumor morphology (aggregated cell pattern, round cell shape, and spindle cell shape) on $F_1$-Score with 'team' as a random effect for the intercept and aggregated cell patterns serving as baseline reference category for comparisons. The fit of the model was determined using the Restricted Maximum Likelihood (REML) method, with a residual variance of $0.053$ and a \ac{BIC} of $-59.3$. Only spindle cell tumors gave significantly different results. }
\label{tab:tumorgroup}
\end{center}
\end{table}

\begin{table}
\begin{center}
\resizebox{\linewidth}{!}{%
\begin{tabular}{lrrrrrr}
\hline
                  &  coef. & std.err. &       z & P$> |$z$|$ & [0.025 & 0.975]  \\
\hline
intercept (30 cases)         &  0.625 &    0.021 & 30.386 &       0.000 &  0.584 &  0.665  \\
feline (20 cases) & -0.063 &    0.018 & -3.576 &       0.000 & -0.098 & -0.029  \\
human  (50 cases) & -0.004 &    0.014 & -0.281 &       0.779 & -0.032 &  0.024  \\
group var         &  0.004 &    0.008 &        &             &        &         \\
\hline
\end{tabular}}
\caption{Results from the linear mixed effects model assessing the influence of species on $F_1$-Score with 'team' as a random effect for the intercept and canine samples (with a support of 30 cases) serving as baseline reference category for comparisons. The fit of the model was determined using the Restricted Maximum Likelihood (REML) method, with a residual variance of $0.053$ and a \ac{BIC} of $-71.94$.}
\label{tab:species}
\end{center}
\end{table}

\begin{table}
\begin{center}
\resizebox{\linewidth}{!}{%
\begin{tabular}{lrrrrrr}
\hline
                  &  coef. & std.err. &       z & P$> |$z$|$ & [0.025 & 0.975]  \\
\hline
intercept (50 cases)      &  0.600 &    0.019 & 31.109 &       0.000 &  0.562 &  0.638  \\
Hamamatsu S360 (30 cases) &  0.066 &    0.014 &  4.700 &       0.000 &  0.038 &  0.093  \\
Hamamatsu S60 (20 cases)  & -0.047 &    0.016 & -2.942 &       0.003 & -0.079 & -0.016  \\
group var       &  0.004 &    0.008 &        &             &        &         \\

\hline
\end{tabular}}
\caption{Results from the linear mixed effects model assessing the influence of scanner on $F_1$-Score with 'team' as a random effect for the intercept and slides scanned with the 3DHistech Pannoramic Scan 2 (3DH, 50 cases) serving as baseline reference category for comparisons. The fit of the model was determined using the Restricted Maximum Likelihood (REML) method, with a residual variance of $0.051$ and a \ac{BIC} of $-100.08$.}
\label{tab:scanner}
\end{center}
\end{table}

\begin{table}
\resizebox{\linewidth}{!}{%
\begin{tabular}{llr}
\hline
category & subcategory & percentage \\
\hline
  \multirow{5}{*}{borderline morphology} & prophase with resemblance to apoptosis & 31.23 \%\\
  & out of focus (scan artefact or thick tissue section) & 27.54 \% \\
   & poor image tissue quality (such as overstained) & 5.67 \%\\
   & not further classified & 23.13 \% \\
   & total & 87.58 \% \\
\hline
 \multirow{3}{*}{ borderline to non mitotic phases} & prophase with early membrane changes &  4.05 \% \\
  & late telophase (with formation of nuclear membrane)& 0.90 \%\\
  &  total & 4.95 \%\\
\hline
 untypical MF morphology &  & 5.85 \%\\
 cut off at image border &  & 0.54 \% \\
 overlooked (clear mitotic figure) &  & 1.08 \% \\
\hline
\end{tabular} }
\caption{Breakdown of mitotic figures that were additionally identified using the phosphohistone H3 (PHH3) stain.}
\label{tab:phh3_breakdown}
\end{table}

\section{Discussion}
\subsection{  Assessment of employed strategies}
The MIDOG 2022 challenge was the first to assess \ac{MF} recognition across multiple tumor types. This extends the range of covariate shifts to the visual context of the \acp{MF}. In the previous challenge, the main domain shift could be attributed to changes in color, sharpness, and depth of field (caused by the differing scanners). In this challenge, the generalization to different tumor types and hence unknown tissue types that surround the \acp{MF} is harder to reflect in dedicated domain generalization strategies, e.g., domain augmentation. This may explain why the participants of this iteration of the challenge did not opt to formulate novel augmentation strategies. It is noteworthy that the top three approaches to track 1 of the challenge used distinctively different strategies to address the pattern recognition problem (semantic segmentation followed by connected components analysis \citep{jahanifar2022stain}, object detection \citep{kotte2022deep} and classification on a sliding window \citep{lafarge2022fine}), highlighting that the \textit{how} (i.e., augmentation, sampling scheme, post-processing) of training was likely more important than the \textit{what} (i.e., the neural network architecture). One commonality between the three top performing approaches to track 1 was that they all used some form of ensembling technique, which has been reported as a strong determinant of success in biomedical challenges \citep{eisenmann2023winner}, and likely contributed directly to domain robustness. 
{ Of note, in contrast to our expectations, approaches in track 2 (using additional data) of our challenge did not show improved performance compared to track 1. This is particularly interesting, given the similarity of the approaches of the TIA Centre team for both tracks. We attribute this to the fact that the additional dataset that was used for training (TUPAC16, \cite{veta2019predicting}) in the approach by \cite{jahanifar2022stain} might introduce a semantic diversity in the labels for mitotic figures, since it used a different annotation process \citep{bertram2020pathologist}. Given the size of the MIDOG22 training set in comparison to the TUPAC16 set, and that the MIDOG22 data set already includes three different scanner-domains for breast cancer (and the TUPAC16 providing another two), there might also be little added informational value by the additional dataset.}

\subsection{  Domain generalization on the test set}
{ 
  Our test set contained three notable conditions that were not part of any training set. First of all, it introduced a new scanner (Hamamatsu S60), which also coincided with specimens from another, yet unseen lab. As the analysis in Table~\ref{tab:scanner} shows, we found a moderate, but significant reduction in performance on samples from this scanner/lab. Secondly, the test set introduced a new species (felines). The regression analysis in Table~\ref{tab:species} yielded only a moderate but significant performance drop for the unseen species. The third condition that was not part of the training set was the group of spindle cell tumors, where we also found a moderate but statistically significant reduction of performance (see Table~\ref{tab:tumorgroup}). It is worth noting that feline soft tissue sarcoma, a tumor with spindle cell morphology, had the lowest scores across individual domains (refer to Table~\ref{tab:tumordomain}).  This is likely to have an impact on both aggregated evaluations. While our test set represents the broadest span of tumor classifications yet, the pool of conditions with shared characteristics remain limited. This prohibits a definitive stratification of influencing aspects at this point.  However, it is worth highlighting that all three domain-defining factors that were new in the test set of the challenge yielded  drops in performance.

It should be noted that the evaluation using $F_1$/AP scores on individual images, and not the collective, tends to skew results towards lower values. Consider the scenario of a tumor image with a solitary \ac{MF}. If this singular cellular event goes undetected (false negative), the $F_1$ score drops from 1.0 to 0.0. On the other hand, if there is a single false positive misdetection, the $F_1$ score reduces from 1.0 to 0.5.
 Now contrast this scenario in the context of an image with 100 \acp{MF}. In this case, the impact of either events changes the scores only marginally. This means, that especially for cases with a low true \acp{MF} count, the $F_1$ score is strongly influenced by the number of false positives. Consequently, we anticipate a more significant decline in the $F_1$ score for low-grade cases, inherently weighting the deviation from the overall expected performance higher in the macro-average. The bootstrapped results, on the other hand, mitigate this problem by aggregating over a larger sample before calculating the metrics.
}

\subsection{  Container-based submission}
The use of containers for the algorithmic submission comes with an increased risk of unintended and unexpected technical failures for the participants. For this reason, we made an independent \textit{preliminary} test set available to the participants. { Additionally, test cases were shipped with the container. }To avoid overfitting of hyperparameters to this set by the participants and, at the same time, to reduce the computational budget required to evaluate the containers, one daily execution was admitted during a two-week time frame prior to the submission. Since overfitting could still not be ruled out, in this version of the challenge, we opted to use four independent (disjointed from the challenge test set) domains in this phase. 

{  
For every detected object, the submission format provided a field for the detection class (\ac{MF} or \ac{NMF}) as well as for the detection score, which we used for an automatic evaluation of the \ac{AP} score. It was not mandatory to provide meaningful values in the score field, however. Consequently, we were only able to determine \ac{AP} scores as well as precision-recall curves for the approaches where these scores were meaningful, whereas the other approaches were excluded in Figures \ref{fig:ap_metrics}, \ref{fig:pr_curve}, and \ref{fig:phh3-gt}. 

Overall, the usage of containers for algorithmic submissions presented challenges that required careful handling, and our approach of providing a preliminary test set and imposing limitations on executions helped address some of the associated risks. However, since the independence of the test set is of utmost importance for a data challenge, we contend that these additional efforts are well-justified.
}


\subsection{  PHH3-assisted ground truth}
The post-challenge evaluation on the alternative, \ac{PHH3}-assisted ground truth yielded overall lower recall values for all approaches. We attribute this to the inclusion of multiple \acp{MF} having inconclusive morphological features in the \ac{HE} image, which could be identified with higher confidence in the \ac{IHC} due to immunopositivity against \ac{PHH3}-antibodies. Equivocal or inconclusive morphologies include the \ac{MF} being out of focus due to the factual three-dimensionality of the sample as well as general difficulty in clearly differentiating some \ac{MF} morphologies (particularly prometaphase \ac{MF}) from imposters. In the \ac{PHH3} stain, however, these structures are clearly distinguishable due to immunoreactivity, which provides an unaltered high contrast, contributing to the overall higher number of \acp{MF}. Similar to the expert annotators of the \ac{HE}-approach, algorithms were trained (based on the ground truth used) to exclude these morphologically inconclusive structures, which explains the lower recall values of all approaches. The good agreement of the challenge ground truth (three-expert { majority vote}) compared to the alternative and \ac{IHC}-assisted ground truth highlights the benefits of multiple blinded expert ensembles for \ac{HE}-based \ac{MF} annotations. The in-depth evaluation of mitotic figures that were only identifiable using the \ac{IHC} stain as secondary source of information (Table~\ref{tab:phh3_breakdown}), however, also reveals the limitations of purely \ac{HE}-based ground truth definitions, as occurring borderline morphological patterns were found to represent the majority of \ac{IHC}-positive \acp{MF} that were not found in the annotations { based solely on the \ac{HE} stain}. { Nonetheless, it is worth pointing out that the \ac{PHH3}-assisted ground truth should not be simply perceived as an improved version of the \ac{HE}-based ground truth on \ac{HE}-stained images, even if it is a more accurate description of the biological truth. The majority of cells that were additionally attributed to be \acp{MF} after consulting the \ac{PHH3} stain were not sufficiently discriminable using the \ac{HE} stain alone since the necessary information was likely just not contained in the images. Training on such annotations could hence exhibit a higher degree of label noise (if only the information available from the \ac{HE} image is considered as input to the network). 

Our analysis indicates that by using the \ac{PHH3}-assisted identification of \ac{MF} objects, we can expect a considerable increase of the \ac{MC}, which is in line with findings by other works on \ac{PHH3} alone \citep{van2020assessment,dessauvagie2015validation}. Currently, grading schemes predominantly rely on \ac{HE}-based counts alone and changing the methodology could invalidate the respective cutoff values.  Hence, the use of \ac{PHH3}-assisted labels for training \ac{MF} detectors could additionally also lead to an overall increase of the \ac{MC}, which would be conflicting with current grading schemes. }

\subsection{  Limitations of the AP metric}
One insight from our challenge is the limitations of the \ac{AP} metric, which averages the precision at defined recall values, as a ranking metric. Besides a high number of hyperparameters (such as the maximum number of detections, the interpolation method, and grid), the \ac{AP} metric is used according to multiple different definitions \citep{hirling2023segmentation}. Moreover, as can be seen in Fig. \ref{fig:pr_curve}, none of the algorithms reached the zero value for precision, which penalized the approaches in the \ac{AP} metric. We hypothesize that this is a result of all approaches using a detection threshold before the non-maximum suppression; a common procedure to reduce computational overhead for the matching of ground truth and candidates, which is an operation in $\mathcal{O}(n^2)$. If no value can be meaningfully interpolated for high recall values (e.g., for the \ac{MIDOG} 2021 baseline approach in Fig. \ref{fig:pr_curve} above a recall value of 0.6), the precision value is commonly extrapolated to 0, which penalizes the approach unjustly. Similarly, should the averaging be confined to the maximum achieved recall value, methods employing a high detection threshold would gain an unfair advantage. In particular, this is demonstrated when comparing the winning approach of \cite{jahanifar2022stain} and the runner-up of \citep{kotte2022deep}. While the precision-recall curve in Fig. \ref{fig:pr_curve} clearly indicates the superiority of the winning approach, the AP metric (see Fig. \ref{fig:ap_metrics}) benefits from the lower detection threshold of the approach by \cite{kotte2022deep}, giving a false impression that the latter approach has a higher decision-threshold independent performance. This provides additional evidence for the utility of the $F_1$ score as the primary challenge metric.

\subsection{  Performance comparison and outlook}

We found that the top algorithmic solutions of this challenge detected \acp{MF} at a level similar to that of the 2021 \ac{MIDOG} challenge (top $F_1$ value of 0.748 in 2021 \citep{aubreville2023mitosis} and 0.764 in 2022). Additionally, comparing these performances to published $F_1$ values for human experts (0.563 for human breast cancer \citep{aubreville2023mitosis}, 0.79 on canine cutaneous mast cell tumor \citep{Bertram2021VetPathol}) indicates that the automatic approaches are in the range of human experts. Nevertheless, it is worth pointing out that human experts typically perform this task not only on \acp{ROI} but on the entire slide, which was not the task of this challenge. We hence encourage the creation of further datasets and challenges incorporating annotations on the entire \acp{WSI} and thus also providing labels for a much more diverse set of tissue characteristics.

\section*{Acknowledgement}
Computational resources and additional support for the challenge have been provided by grand-challenge.org. The challenge organizers would like to thank Siemens Healthineers and Tribun Health for donating the monetary prizes of the challenge, which have been awarded to the top three participants in { the first track and the two teams in the second track}. The organizers would further like to thank Medical Data Donors e.V. for providing assistance in the organization of the awards. J.A. acknowledges support from the Bavarian Institute for Digital Transformation (project ReGInA). M.A. and R.K. acknowledge support by the German Research Foundation (project number 520330054). K.B. and F.W. received funding by the German Research Foundation (DFG) project 460333672 CRC1540 EBM. K.B. further acknowledges support by d.hip campus in form of a faculty endowment.  C.A.B. acknowledges funding by the Austrian Science Fund (FWF, project number: I 6555). 

\section{Author contributions}
The challenge was organized by Marc Aubreville, Katharina Breininger, Frauke Wilm, Christof A. Bertram, Samir Jabari, Nikolas Stathonikos, and Mitko Veta.

Frauke Wilm, Jonas Ammeling, and Jonathan Ganz provided the algorithmic reference approaches for the challenge.

The evaluations for this paper were carried out by Marc Aubreville. Christof A. Bertram and Marc Aubreville wrote the main text of this work. Taryn A. Donovan, as a native speaker, has made additional language corrections. All authors reviewed the manuscript. { Jonas Ammeling and Marc Aubreville did the statistical evaluation. }

Christof A. Bertram, Robert Klopfleisch, and Taryn A. Donovan served as expert pathologists in annotating the complete challenge dataset. Nikolas Stathonikos, Robert Klopfleisch, Samir Jabari, Christof Bertram, and Markus Eckstein provided samples for the challenge.

Jonas Annuscheit and Christian Krumnow (Team HTW Berlin), Engin Bozaba (Team Virasoft), Mostafa Jahanifar and Adam Shephard (Team TIA Centre), Satoshi Kondo and Satoshi Kasai (Team SKJP), Sujatha Kotte and VG Saipradeep (Team TCS Research), Maxime W. Lafarge and Viktor H. Koelzer (Team USZ / UZH Zurich), Ziyue Wang and Yongbing Zhang (Team HITszCPATH), Sen Yang and Xiyue Wang (Team AI\_medical) were participants of the challenge. All participants contributed to the overview of the submitted methods section.

\bibliographystyle{plainnat}
\bibliography{refs}

\clearpage
\appendix
\section{Supplementary Material}

\begin{figure*}[ht!]
\includegraphics[width=0.9\textwidth]{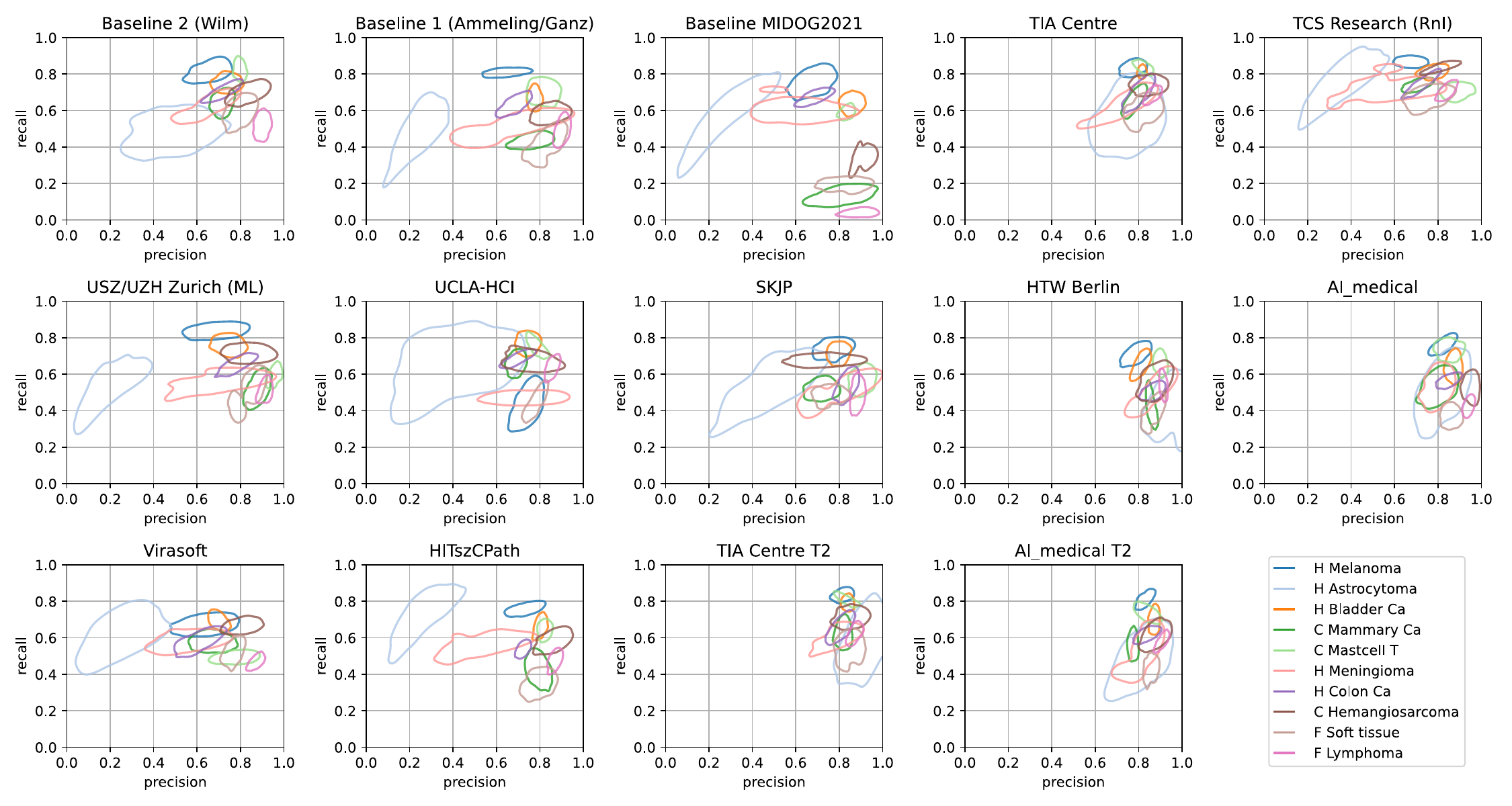}
\caption{80\% confidence regions for precision and recall for each team and tumor type, plotted for each individual team. Confidence intervals were established using a Gaussian kernel density estimator applied to empirically bootstrapped datasets.}
\end{figure*}

\begin{figure}[h!]
\centering
\includegraphics[width=0.6\linewidth]{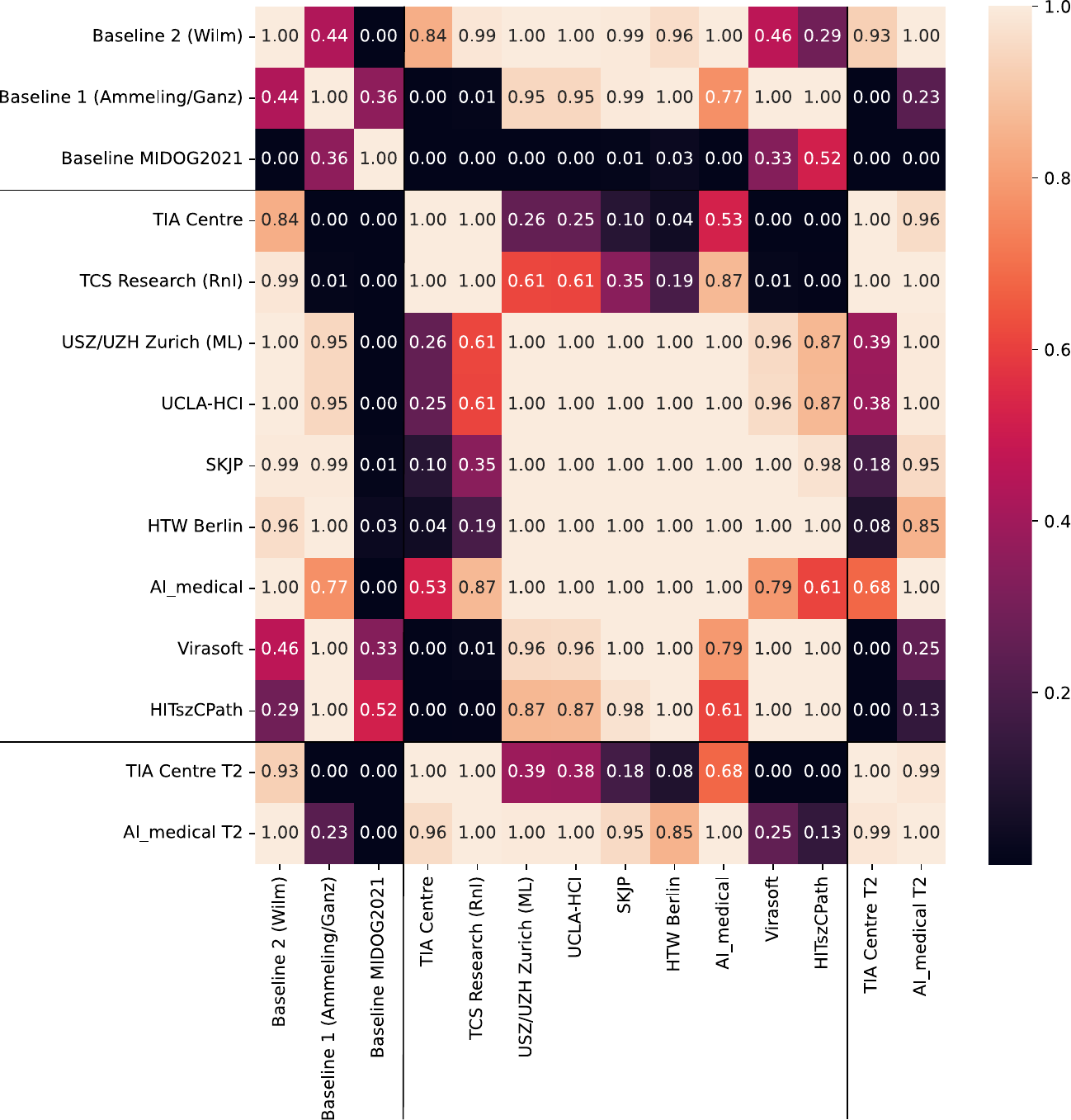}
\caption{Results of the Tukey HSD test, assessing results of all approaches for statistical significance. Table shows p values.}
\label{fig:tukey_significance}
\end{figure}



\begin{acronym}
\acro{ROI}[ROI]{regions of interest}
\acro{MIDOG}[MIDOG]{MItosis DOmain Generalization}
\acro{HE}[H\&E]{hematoxylin and eosin}
\acro{ICPR}[ICPR]{the International Conference on Pattern Recognition}
\acro{MICCAI}[MICCAI]{Medical Image Computing and Computer Assisted Intervention}
\acro{MF}[MF]{mitotic figure}
\acro{NMF}[NMF]{non-mitotic figure}
\acro{PHH3}[PHH3]{Phospho-Histone H3}
\acro{IHC}[IHC]{immunohistochemical}
\acro{MC}[MC]{mitotic count}
\acro{DA}[DA]{diagnostic archive}
\acro{TTA}[TTA]{test-time augmentation}
\acro{DETR}[DETR]{Detection Transformer}
\acro{UMC}[UMC]{University Medical Center}
\acro{FUB}[FUB]{Freie Universität Berlin}
\acro{CNN}[CNN]{convolutional neural network}
\acro{WSI}[WSI]{whole slide image}
\acro{VMU}[VMU]{University of Veterinary Medicine Vienna}
\acro{AP}[AP]{average precision}
\acro{BIC}[BIC]{Bayesian information criterion}
\end{acronym}

\end{document}